%% file: acl_latex.tex
\documentclass[11pt]{article}

\usepackage[preprint]{acl}

\usepackage{times}
\usepackage{latexsym}

\usepackage[T1]{fontenc}

\usepackage[utf8]{inputenc}

\usepackage{microtype}

\usepackage{inconsolata}

\usepackage{graphicx}

\usepackage{hyperref}
\usepackage{url}
\usepackage{booktabs}       
\usepackage{amsfonts}       
\usepackage{subfigure}     
\usepackage{nicefrac}       
\usepackage[table]{xcolor}  
\usepackage{tablefootnote}  
\usepackage[flushleft]{threeparttable}  
\usepackage{multirow}
\usepackage{amsmath}
\usepackage{tcolorbox}
\usepackage{listings}
\usepackage{minitoc} 
\usepackage[page,header]{appendix}
\usepackage{titletoc}
\usepackage{colortbl}      

\title{xVerify: Efficient Answer Verifier for Reasoning Model Evaluations}

\author{
  Ding Chen\textsuperscript{\rm 1}\thanks{Equal contribution. $\dagger$ Corresponding authors} \quad
  Qingchen Yu\textsuperscript{\rm 2}\footnotemark[1] \quad
  Pengyuan Wang\textsuperscript{\rm 2}\footnotemark[1] \quad
  Mengting Hu\textsuperscript{\rm 3} \quad
  \\
  \textbf{Wentao Zhang\textsuperscript{\rm 4}\footnotemark[2]} \quad
  \textbf{Zhengren Wang\textsuperscript{\rm 5}} \quad
  \textbf{Bo Tang\textsuperscript{\rm 2}} \quad
  \textbf{Feiyu Xiong\textsuperscript{\rm 2}} \quad
  \\
  \textbf{Xinchi Li\textsuperscript{\rm 1}} \quad
  \textbf{Chao Wang\textsuperscript{\rm 6}} \quad
  \textbf{Mingchuan Yang\textsuperscript{\rm 1}} \quad
  \textbf{Zhiyu Li\textsuperscript{\rm 2}\footnotemark[2]}\\
  \\
  \textsuperscript{\rm 1} China Telecom Research Institute \quad
  \textsuperscript{\rm 2} MemTensor (Shanghai) Technology Co., Ltd. \\
  \textsuperscript{\rm 2} College of Software, Nankai University \quad
  \textsuperscript{\rm 4} Center for Data Science, Peking University \\
  \textsuperscript{\rm 5} Peking University \quad
  \textsuperscript{\rm 6} Data Development Center of China Telecom \\
  \texttt{wentao.zhang@pku.edu.cn, lizy@iaar.ac.cn}
}

\begin{document}
\maketitle
\begin{abstract}
With the release of OpenAI’s o1 model, reasoning models that adopt slow-thinking strategies have become increasingly common. Their outputs often contain complex reasoning, intermediate steps, and self-reflection, making existing evaluation methods and reward models inadequate. In particular, they struggle to judge answer equivalence and to reliably extract final answers from long, complex responses. To address this challenge, we propose xVerify, an efficient answer verifier for evaluating reasoning models. xVerify shows strong equivalence judgment capabilities, enabling accurate comparison between model outputs and reference answers across diverse question types. To train and evaluate xVerify, we construct the VAR dataset, which consists of question–answer pairs generated by multiple LLMs across various datasets. The dataset incorporates multiple reasoning models and challenging evaluation sets specifically designed for reasoning assessment, with a multi-round annotation process to ensure label quality. Based on VAR, we train xVerify models at different scales. Experimental results on both test and generalization sets show that all xVerify variants achieve over 95\% F1 score and accuracy. Notably, the smallest model, xVerify-0.5B-I, outperforms all evaluation methods except GPT-4o, while xVerify-3B-Ib surpasses GPT-4o in overall performance. In addition, reinforcement learning experiments using xVerify as the reward model yield an 18.4\% improvement for Qwen2.5-7B compared with direct generation, exceeding the gains achieved with Math Verify as the reward. These results demonstrate the effectiveness and generalizability of xVerify. All xVerify resources are available on \href{https://github.com/IAAR-Shanghai/xVerify}{GitHub}.
\end{abstract}

\section{Introduction}

With the emergence of chain of thought (CoT) prompting~\citep{cot_2022_nips_google}, researchers began to explicitly encourage LLMs to generate intermediate reasoning steps, thereby enhancing their ability to handle complex tasks. Following this, OpenAI introduced the o1 model~\citep{o1_2024_arXiv_openai}, which proposed the concepts of slow thinking and scaling at test time. Specifically, the model is trained to output a detailed reasoning process before generating a final answer, significantly improving its performance on complex tasks. Inspired by this paradigm, a variety of reasoning models have emerged, such as DeepSeek-R1~\citep{deepseek_r1_25_deepseek} trained with GRPO, OpenAI's o3-mini~\citep{o3_mini_25_openai}, and QwQ-32B~\citep{qwq32b_25_qwen}. However, the rise of reasoning models has posed substantial challenges for evaluation. Because the reasoning processes output by these models may include redundant information, intermediate results, and self-reflections, current evaluation methods, as well as reward models in reinforcement learning, often prove ineffective~\citep{survey_2024_arXiv_Microsoft}.

Developing scalable and accurate evaluation methods for LLMs on complex reasoning tasks (e.g., commonsense, logical, multi-hop, and mathematical reasoning) has become increasingly important~\citep{eval_survey_2023_arXiv_tianjin}. While human annotation remains the gold standard, it is labor-intensive and difficult to scale. Automatic methods fall into two main categories: rule-based frameworks~\citep{opencompass_24_github, ultraeval_24_Tsinghua, math_verify_24_github, evals_24_github_OPENAI}, which extract answers through strict formatting and pattern matching, and LLM-based judge models~\citep{llm_judge_survey_2024_arXiv_Tsinghua,llm_judge_survey_2025_arXiv_IDEA,llm_judge_survey_2025_arXiv_Arizona}, which provide qualitative assessments or scores~\citep{survey_2024_arXiv_Microsoft}. Both types of methods are commonly used to evaluate the performance of final models or to serve as reward models in reinforcement learning. However, rule-based methods struggle with diverse output formats and lengthy chains of thought; for example, Math-Verify~\citep{math_verify_24_github}, adopted in the Open-R1\footnote{\href{https://github.com/huggingface/open-r1}{https://github.com/huggingface/open-r1}} project, can only handle mathematical results that are strictly formatted and in fixed positions. Although judge models offer adaptability, they are not explicitly trained for objective 'correct/incorrect' decision-making~\citep{llm_judge_survey_2025_arXiv_IDEA}. Consequently, a robust, automated solution specifically tailored for objective reasoning evaluation is still lacking.

To address these challenges, we introduce xVerify, an efficient LLM-answer verifier tailored for evaluating LLM responses to objective questions. xVerify processes the full LLM output, enabling it to accurately identify final answers from complex reasoning traces. It also supports robust equivalence checking, including symbol conversion (e.g., alpha → $\alpha$), mathematical expression matching, and semantic alignment in natural language. Moreover, it is tolerant of formatting errors such as malformed LaTeX, making it applicable to a wide range of tasks, including math problems, multiple-choice, short-answer, and classification questions. To train and evaluate xVerify, we construct the \textbf{V}erify \textbf{A}nswer for \textbf{R}easoning (VAR) dataset, which includes responses from 19 LLMs across 24 reasoning benchmarks. All labels are verified through multi-round GPT-4o and human review. The dataset covers advanced reasoning models and benchmarks like GPQA, LiveMathBench, and AIME 2024. We fine-tune xVerify on a variety of base models (e.g., Qwen2.5, LLaMA, Gemma 2) and scales (0.5B–32B). Remarkably, even the smallest variant (xVerify-0.5B-I) surpasses existing evaluation methods—including 32B-sized models—on all metrics, while larger variants achieve F1 and accuracy over 95\% on both test and generalization sets. Furthermore, we conduct reinforcement learning (RL) experiments with xVerify as the reward model. Compared with direct generation, it shows an improvement of 18.4\% for Qwen2.5-7B. This also represented a greater improvement than when Math Verify is used as the reward model. For Llama3.1-8B, we achieve similar improvements.

The main contributions of this paper can be summarized in three key points:
\begin{itemize}
  \item We construct the VAR dataset, which contains answer samples from 19 LLMs across 24 evaluation benchmarks. The dataset is annotated via multiple rounds of GPT-4o and human review, and is designed for training and evaluating judge models for reasoning tasks.
  \item We propose xVerify, an efficient answer verifier for evaluating reasoning models, and have released several fine-tuned versions that are publicly available on Hugging Face.
  \item We comprehensively evaluate xVerify in two key capacities: as a judge model, demonstrating superior accuracy and robustness against existing methods on both in-domain and out-of-distribution benchmarks; and as a reward model in RL, where it improves policy performance over direct generation.
\end{itemize}

\section{Related Work}

Evaluation methods are a crucial component in the development of LLM~\citep{survey_2024_arXiv_Microsoft}. However, the open-ended nature of LLM outputs makes it difficult to apply standardized metrics, limiting the effectiveness of traditional evaluation methods~\citep{llm_judge_survey_2024_arXiv_Tsinghua}. The rise of reasoning models~\citep{o3_mini_25_openai,deepseek_r1_25_deepseek,qwq32b_25_qwen}, generating lengthy and complex reasoning, further complicates evaluation. For objective tasks, the main challenge is to accurately extract the final answer from the LLM's semi-structured output and compare it with the reference answer. Existing approaches are typically divided into human evaluation and automatic evaluation. While human evaluation offers flexibility, automatic methods are more cost-efficient and consistent~\citep{survey_2024_arXiv_Microsoft}. Current automatic methods mainly include rule-based evaluation frameworks and LLM-based judgment methods.

Rule-based methods are widely used in automatic evaluation frameworks such as LM Eval Harness~\citep{lm_eval_harness_21_Zenodo}, OpenCompass~\citep{opencompass_24_github}, UltraEval~\citep{ultraeval_24_Tsinghua}, and OpenAI Evals~\citep{evals_24_github_OPENAI}. Tools like Math-Verify~\citep{math_verify_24_github} also follow this approach, extracting final answers using regular expressions (RegEx) and comparing them with reference answers. However, LLM outputs often contain final answers in varied surface forms—e.g., "alpha" vs. "$\alpha$", "A" vs. "a", or "1000" vs. "$10^3$"—which can be semantically equivalent but textually different. While some tools support limited transformations, they typically handle only LaTeX expressions or simple string patterns, and struggle with basic semantic equivalence like "one hundred" vs. "100". For reasoning models, the output is usually lengthy and involves complex reasoning steps with intermediate results. This makes it difficult for regular expressions to accurately identify the final answer, causing rule-based approaches to frequently fail in such contexts. Moreover, prior work has shown that LLMs may revise or overturn their initial predictions during extended reasoning processes, exhibiting a kind of self-reflection~\citep{FTEvsTOE_24_arxiv_LMU}. Additionally, rule-based methods typically ignore the reasoning process and only evaluate the final answer, which has drawn criticism from many researchers—especially in the context of reasoning models~\citep{cotper_2022_nips_google,self_consistency_2023_iclr_google,chain_2024_acl_google}. Thus, rule-based evaluations are limited in reasoning scenarios.

LLM-based judgment methods use fine-tuned LLMs to evaluate the quality of other LLMs’ responses. Compared to traditional evaluation methods, they offer greater task adaptability, generate interpretable results, reduce evaluation costs, and can be applied across the LLM lifecycle~\citep{llm_judge_survey_2024_arXiv_Tsinghua,llm_judge_survey_2025_arXiv_IDEA,llm_judge_survey_2025_arXiv_Arizona}. For objective questions, these judge models can extract final answers from responses with intermediate reasoning or self-reflection. In recent years, many LLM-based judge models have emerged, including JudgeLM~\citep{judgelm_25_icrl}, PandaLM~\citep{pandalm_24_icrl_Peking}, Auto-J~\citep{auto_j_24_iclr_JiaoTong}, Prometheus 2~\citep{prometheus_2_24_acl_KAIST}, CompassJudger~\citep{CompassJudger_1_24_arXiv_Shanghai_AI}, CritiqueLLM~\citep{critiquellm_24_acl_Tsinghua}, and Themis~\citep{themis_24_emnlp_Peking}. Judge models typically support pointwise, pairwise, and listwise evaluations~\citep{llm_judge_survey_2024_arXiv_Tsinghua}, and some also serve as reward models in reinforcement learning. However, most are designed to assign scores to LLM outputs, making them more suitable for subjective evaluations like helpfulness, reliability, or relevance. For objective questions that require binary decisions (“correct” or “incorrect”), these models are less effective. Although scores can be binarized using thresholds, this approach is unreliable, as the models are not explicitly trained for such tasks. Moreover, the current LLM-based critic models and PRMs (Process Reward Models) exhibit subpar performance when detecting errors in long chain-of-thought responses generated by reasoning models~\citep{deltabench_2025_arXiv_alibaba}. Thus, while judge model holds promise for evaluating reasoning models, they require targeted training.

In summary, automatic evaluation on objective tasks remains underdeveloped. Rule-based and LLM-based methods each have clear limitations, while human annotation is costly and hard to scale. To address these challenges, we propose xVerify, a robust and targeted judge model specifically designed for objective evaluation of LLMs.

\begin{figure*}[!htp]
    \vspace{-1.em}
    \centering
    \includegraphics[width=0.8\linewidth]{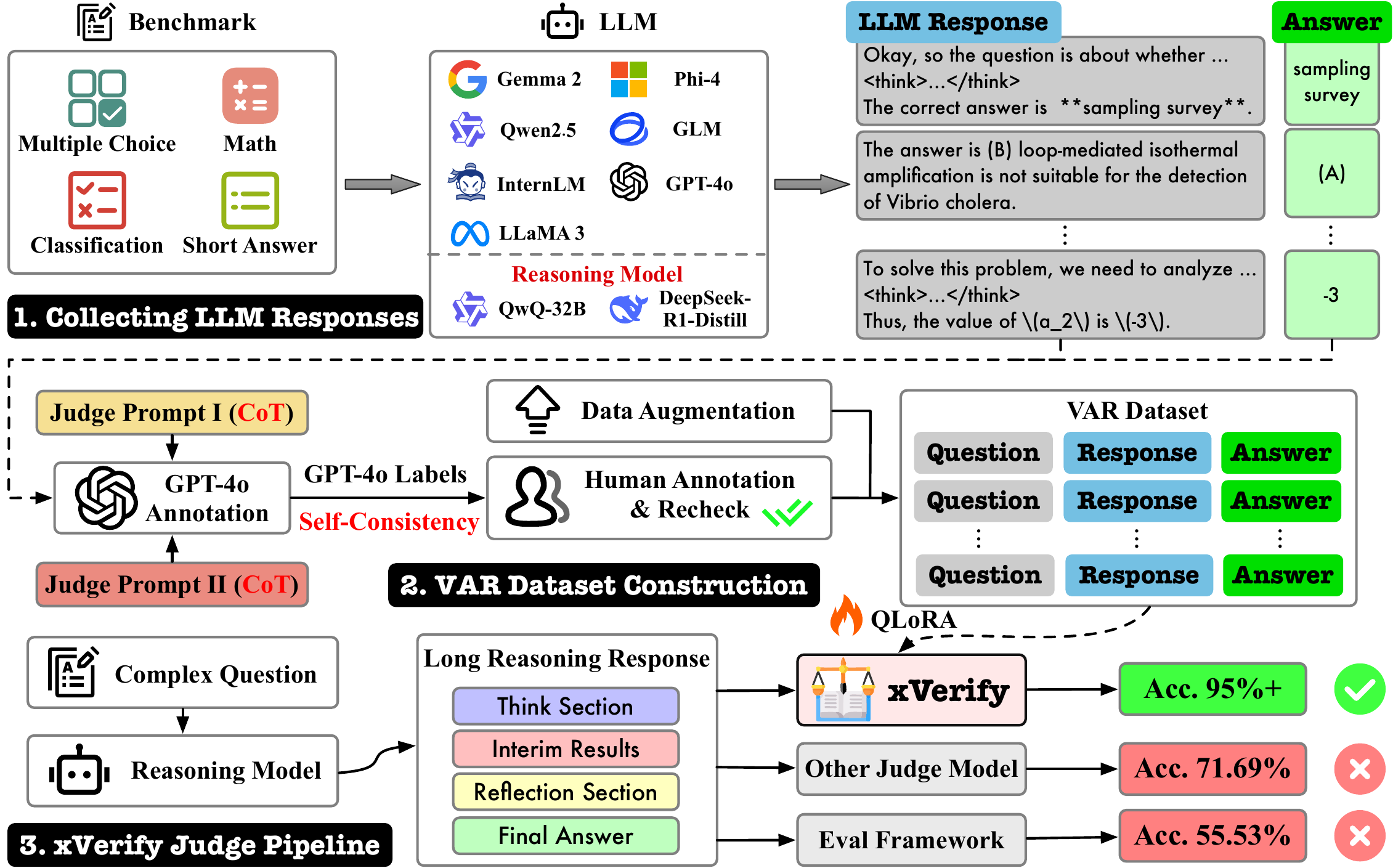}
    \caption{Framework of xVerify: (1) Collecting LLM Responses: aggregate responses from multiple LLMs across datasets covering four question types. (2) VAR Dataset Construction: employ GPT-4o and human annotators for labeling and rechecking, and use data augmentation to refine the dataset. (3) xVerify Judge Pipeline: accurately evaluate multi-component answers from reasoning models on challenging questions.} 
    \label{fig:framework}
    \vspace{-1.em}
\end{figure*}

\section{Problem Definition}
\label{sec:Problem Definition}
To evaluate the correctness of LLM responses to objective questions, the key is to extract the final answer from the response and compare it with the reference answer. We formally define this evaluation task as follows:

We formalize this task as a 4-tuple $(\mathrm{Q}, \mathrm{R}, \mathrm{A_{ref}}, \mathrm{E})$, where $\mathrm{Q} = \{q_1, q_2, ..., q_n\}$ is the set of questions, $\mathrm{R} = \{r_1, r_2, ..., r_n \mid r_i = \mathcal{W}(q_i)\}$ is the set of responses generated by an LLM $\mathcal{W}$, $\mathrm{A_{ref}} = \{a_{ref}^1, ..., a_{ref}^n\}$ is the set of reference answers, and $\mathrm{E}: \mathrm{Q} \times \mathrm{R} \times \mathrm{A_{ref}} \rightarrow {0, 1}$ is the evaluation function that returns 1 if the response is correct and 0 otherwise. 

For the stage of extracting the final answer, given a response $r$ to question $q$, which may include intermediate reasoning and multiple candidate answers, we denote the extracted candidates as $\mathrm{A}(r)$. To identify the final answer, we define a scoring function $\mathrm{S}: \mathrm{A}(r) \times \mathrm{Q} \rightarrow \mathbb{R}$ that measures the relevance or suitability of each candidate $a \in \mathrm{A}(r)$ to $q$, and select the final answer using the extraction function: $\varepsilon(q, r) = \arg\max_{a \in \mathrm{A}(r)} \mathrm{S}(a, q). $

For the equivalence comparison stage, we define an equivalence function $\psi: \mathrm{A_{ref}} \times \mathrm{A_{final}} \rightarrow \{0, 1\}$, where $\psi$ returns 1 if the predicted answer is equivalent to the reference, and 0 otherwise. Since answers may appear in different forms, $\psi$ integrates results from the following three sub-functions:

For mathematical expressions, we define a composite normalization function \(\Phi_{\text{norm}}^{math} = \phi_{\text{err}} \circ \phi_{\text{syn}} \circ \phi_{\text{alg}} \circ \phi_{\text{dim}}\), where $\phi_{\text{err}}$ repairs minor syntax errors, $\phi_{\text{syn}}$ unifies syntactic structures, $\phi_{\text{alg}}$ performs algebraic simplification, and $\phi_{\text{dim}}$ ensures consistency in physical units. By transforming expressions into a canonical form, $\Phi_{\text{norm}}^{math}$ enables reliable equivalence comparison:

{\small
\begin{equation}
\begin{aligned}
    \psi_{math}(a_{ref}^{math}, a_{final}^{math}) 
    = 
    \begin{cases}
    1 & \text{if }\Phi_{\text{norm}}^{math}(a_{ref}^{math})  \\ &= \Phi_{\text{norm}}^{math}(a_{final}^{math}), \\
    0 & \text{otherwise}
    \end{cases}
\end{aligned}
\label{eq:math equal function}
\end{equation}
}

For natural language answers, we define a comparison function \(\psi_{\text{nl}}: \mathrm{A_{ref}^{nl}} \times \mathrm{A_{final}^{nl}} \rightarrow \{0, 1\}\) to assess semantic equivalence. Specifically, we introduce a semantic alignment function \(\phi_{\text{align}}^{nl}\) to measure the similarity between two textual answers. The equivalence decision is made by comparing the alignment score with a predefined threshold \(\tau\):

{\small
\begin{equation}
    \psi_{nl}(a_{ref}^{nl}, a_{final}^{nl}) = 
    \begin{cases}
    1 & \text{if } \phi_{\text{align}}^{nl}(a_{ref}^{nl}, a_{final}^{nl}) \geq \tau, \\
    0 & \text{otherwise}
    \end{cases}
    \label{eq:nl equal function}
\end{equation}
}

For symbolic representations, we define a composite normalization function \(\Phi_{\text{norm}}^{sym} = \phi_{\text{uni}} \circ \phi_{\text{font}} \circ \phi_{\text{dom}}\), which unifies symbols by applying \(\phi_{\text{uni}}\) for Unicode normalization, \(\phi_{\text{font}}\) for aligning font styles, and \(\phi_{\text{dom}}\) for domain-specific mappings. This produces a standardized form for character-level comparison, and the $\Phi_{\text{norm}}^{sym}$ is defined as:

{\small
\begin{equation}
\begin{aligned}
    \psi_{sym}(a_{ref}^{sym}, a_{final}^{sym}) = 
    \begin{cases}
    1 & \text{if }\Phi_{\text{norm}}^{sym}(a_{ref}^{sym}) \\ &= \Phi_{\text{norm}}^{sym}(a_{final}^{sym}), \\
    0 & \text{otherwise}
    \end{cases}
\end{aligned}
\label{eq:symbol equal function}
\end{equation}
}

Based on the above components, we define a unified equivalence function \(\psi\) to determine whether the final answer \(a_{final}\) matches the reference answer \(a_{ref}\) across different modalities. Defined as:

{\small
\begin{equation}
    \psi(a_{final}, a_{ref}) = 
    \begin{cases}
    1, & \text{if } \psi_{math}(a_{final}^{math}, a_{ref}^{math}) = 1 \\
       & \quad \land\ \psi_{nl}(a_{final}^{nl}, a_{ref}^{nl}) = 1 \\
       & \quad \land\ \psi_{sym}(a_{final}^{sym}, a_{ref}^{sym}) = 1, \\
    0, & \text{otherwise}
    \end{cases}
    \label{eq:overall_equiv}
\end{equation}
}

Here, \(a_{final}^{math}\), \(a_{final}^{nl}\), and \(a_{final}^{sym}\) represent the mathematical, natural language, and symbolic parts of the final answer, respectively, and similarly for \(a_{ref}\). This allows for equivalence checking in both unimodal and multimodal settings.

To summarize, the overall evaluation function \(\mathrm{E}\) is defined as: $\mathrm{E}(q, r, a_{ref}) = \psi\big(\varepsilon(q, r),\ a_{ref}\big)$, where \(q\) is the objective question, \(r\) is the response generated by the LLM, and \(a_{ref}\) is the corresponding reference answer.

\section{Methodology}

The xVerify training and evaluation pipeline includes three main stages: collecting LLM responses, VAR dataset construction, and the xVerify judge pipeline (Figure~\ref{fig:framework}). We first gather question–response pairs from various LLMs across four types of objective questions, including complex, reasoning-intensive examples. To ensure accurate labels, we employ multiple rounds of annotation and rechecking using both GPT-4o and human annotators. We also apply data augmentation to increase the dataset's diversity and complexity. Finally, we train xVerify models of different sizes on the VAR dataset to evaluate long, multi-step answers—cases challenging for existing evaluation methods. Section~\ref{sec:var dataset} details the dataset construction, and Section~\ref{sec:model training} describes the training process.

\begin{figure*}[htp]
    \vspace{-1.em}
    \centering
    \includegraphics[width=0.8\linewidth]{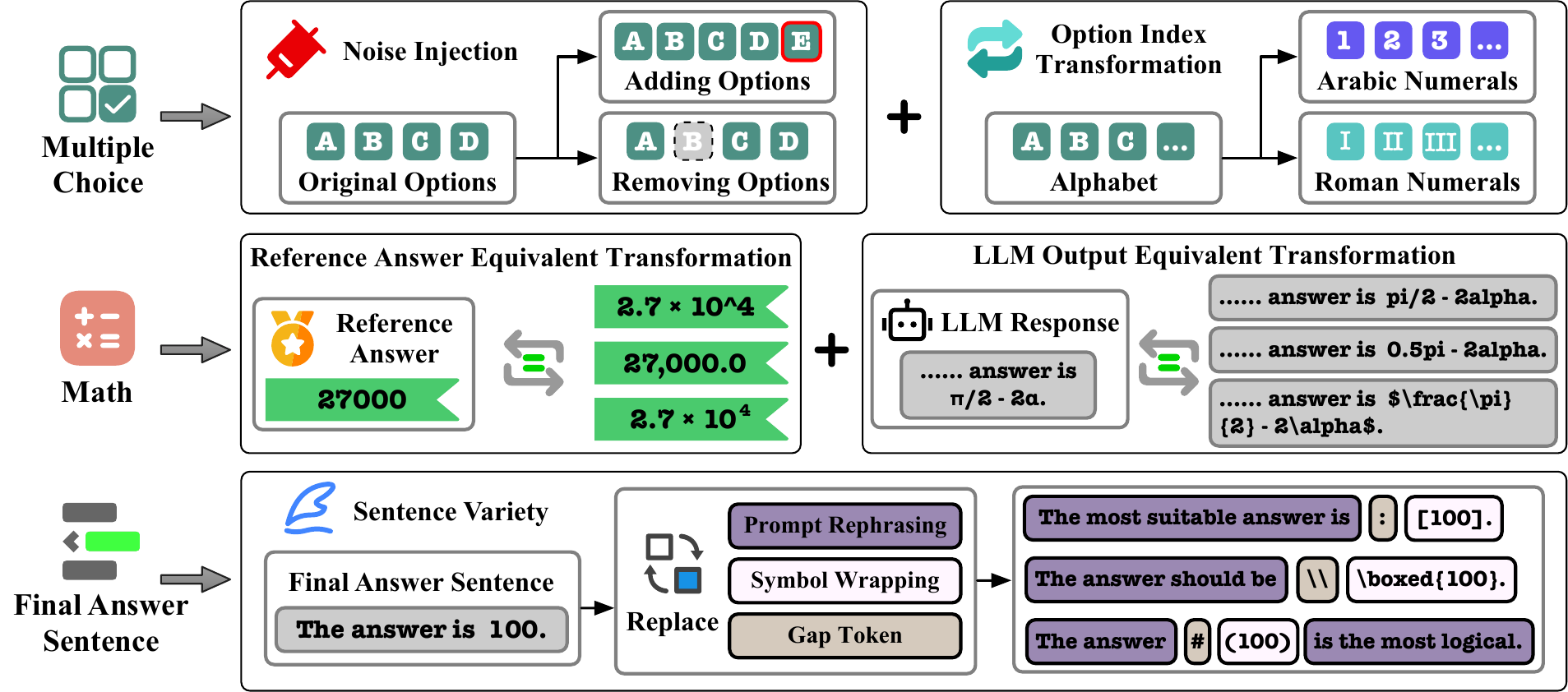}
    \caption{Data Augmentation Pipelines: (1) transformation of multiple-choice options through numbering conversion and noise injection, (2) diversification of mathematical answers via equivalent expression generation, and (3) final answer sentence transformation using prompt rephrasing, symbol wrapping, and gap token insertion.} 
    \label{fig:data augmentation framework}
    \vspace{-1.em}
\end{figure*}

\subsection{VAR Dataset}
\label{sec:var dataset}

xVerify is designed to assess the correctness of reasoning models’ responses on objective questions. However, current judge models are mostly trained on tasks such as scoring or reviewing, and reasoning models with lengthy responses have only recently emerged. Thus, no suitable dataset exists for training xVerify. To better train and evaluate xVerify, we constructed a dedicated dataset named Verify Answer for Reasoning (VAR). Examples from the VAR dataset are provided in Appendix~\ref{appendix:Examples from the VAR Dataset}.

\subsubsection{LLM Response Generation}

To ensure the diversity and coverage of the dataset, we selected 19 mainstream LLMs and 24 frequently used multilingual datasets to generate and collect responses. To better simulate the answering patterns of reasoning models in common evaluation scenarios, the chosen LLMs include recent models such as the DeepSeek-R1-Distill series~\citep{deepseek_r1_25_deepseek} and QwQ-32B~\citep{qwq32b_25_qwen}. Most of the other LLMs also support context lengths exceeding 32k tokens, enabling them to produce answers with extended reasoning chains. The selected datasets include high-difficulty benchmarks commonly used for evaluating reasoning models, such as GPQA~\citep{gpqa_2024_colm_New_York}, AIME 2024~\citep{AIME_2024}, MATH~\citep{math_2021_nips_UC}, and LiveCodeBench~\citep{livemathbench_2024_arXiv_shanghailab}, which typically require multi-step reasoning and computation to solve. During data generation, we also retained some extremely long responses, such as those exceeding 6k characters in length. Details on all LLMs and datasets are in Appendix~\ref{appendix:datasets and models}.

To train and evaluate xVerify more effectively, we grouped the 24 datasets into four types based on question and answer formats: multiple choice, math, short answer, and classification. Multiple choice questions offer several labeled options; math includes questions where answers are mathematical expressions (e.g., numbers, equations) in mathematics and physics; short answer questions expect brief natural language responses like names or dates, with no strict format constraints; classification tasks involve selecting the correct label, such as for sentiment or topic classification.

To reflect realistic evaluation settings and generate a diverse set of Q\&A samples, we designed multiple prompt templates for guiding the LLMs in response generation. The prompt configurations vary along several dimensions: 0-shot vs. 5-shot, with or without CoT, and with or without answer format restrictions (restrict), resulting in eight distinct prompt types. Details of all prompt templates are provided in Appendix~\ref{appendix:prompts for generating LLM responses}.

In total, we generated 191,600 Q\&A samples using the 19 LLMs and 24 evaluation sets, providing a diverse sample pool for constructing the dataset.

\subsubsection{Dataset Partitioning}

Based on the previously collected sample pool, we constructed the training, test, and generalization sets through filtering and preprocessing.

The training and test sets are used to train and evaluate xVerify. Both are sampled from the same pool, sharing similar distributions. Specifically, they include samples generated by 15 LLMs across 17 evaluation sets, covering the four question types. The training set contains 36,941 samples, and the test set includes 5,194 samples.

The generalization set complements the test set by evaluating xVerify’s ability to handle more diverse and challenging distributions, reflecting real-world scenarios. It consists of 5,366 samples from 7 evaluation sets not used in the training or test sets, while still spanning all four question types. These samples are generated by 19 LLMs, including 4 models not seen in training or testing, such as the reasoning model QwQ-32B, resulting in greater diversity and distribution shift.

Section~\ref{sec:Data Augmentation} introduces our data augmentation strategy, which adds more challenging samples to all three sets. Detailed dataset statistics are provided in Appendix~\ref{appendix:three set details}.

\subsubsection{Data Annotations}

To ensure the accuracy of xVerify’s training and evaluation, we conducted multiple rounds of automatic and manual annotation across the three datasets. Specifically, we used GPT-4o to perform two rounds of annotation for all samples in the datasets, utilizing two distinct prompt templates (details provided in Appendix~\ref{appendix:prompts for GPT-4o annotation}) to improve annotation confidence~\citep{self_consistency_2023_iclr_google,self_feedback_2024_arXiv_iaar}. Given the large size of the training set, we only applied manual annotation to the more challenging math problems and to samples where the two rounds of GPT-4o annotations disagreed. In contrast, for the test and generalization sets, we manually annotated all samples, resulting in a three-round annotation process to maximize label reliability. Details of the manual annotation process are provided in Appendix~\ref{appendix:Human Annotation}.

\input{tables/test_results}

\subsubsection{Data Augmentation}
\label{sec:Data Augmentation}

To further enhance the diversity and robustness of the dataset, we designed a series of data augmentation strategies (illustrated in Figure~\ref{fig:data augmentation framework}) to better simulate real-world evaluation settings and improve the model’s tolerance to varied answer formats.

For multiple-choice questions, we applied two augmentations: option index transformation and noise injection. The former converts alphabetical labels to Arabic or Roman numerals, while the latter randomly adds or removes irrelevant distractor options without changing the original question intent, thereby increasing structural complexity.

For math problems, we used two approaches: augmentation based on reference answers and LLM responses. In the first approach, we generated 3–5 mathematically equivalent expressions of each reference answer through symbolic and formal transformations, then created new samples accordingly. In the second, we applied the same transformation logic to the final answers in LLM responses, enriching the dataset with varied mathematical formats and helping the model learn equivalence across symbolic expressions.

We also augmented the final answer statements. Specifically, we extracted answer-bearing sentences from responses generated using restrict prompts, and applied over 1,000 transformation patterns. These included: 20 variations of prompt rephrasing (e.g., “The answer is B” → “The most appropriate answer is B”), 18 symbolic wrappers (e.g., wrapping B as \(\boxed{B}\)), and 5 forms of delimiter insertions (e.g., adding a colon or space before the answer). This improved diversity in answer formats and reduced overfitting to specific templates.

Together, these strategies expanded the expressive space of the dataset while preserving semantic consistency, offering richer and more challenging training signals for xVerify. After augmentation, the sizes of the training, test, and generalization sets increased to 43,204, 6,122, and 6,468 samples respectively. Full dataset details are provided in Appendix~\ref{appendix:three set details}. The augmentation of math problems primarily relied on GPT-4o; prompt templates are listed in Appendix~\ref{appendix:prompts for data augmentation}.

\subsection{Model Training}
\label{sec:model training}

We trained 14 models with different parameter sizes and architectures using the training set from the VAR dataset. Specifically, we utilized the LLaMA-Factory framework~\citep{llamafactory_2024_acl_buaa} and QLoRA technique~\citep{dettmers2023qlora} for model training. Based on extensive experimentation, we set the number of epochs to 1 and selected a learning rate of 1e-4 as the optimal configuration, with other hyperparameters detailed in Appendix~\ref{appendix:train hyperparameters}. Many researchers have pointed out potential bias in using LLMs as judge models, where models from the same family tend to receive higher ratings~\citep{leakage_2025_arXiv_arizona}. To thoroughly evaluate the generalization capability of the xVerify method, we trained 14 models with varying parameter sizes and architectures. These models ranged from 0.5B to 32B parameters and included five different families, such as LLaMA 3~\citep{llama3_2024_arXiv_meta}, Qwen2.5~\citep{qwen25_2024_arXiv_qwen}, and Gemma 2~\citep{gemma2_2024_arXiv_google}. Details of the models used are provided in Appendix~\ref{appendix:original model details}.

\section{Experiments}
\label{section:experiments}

This section details our experiments, covering two main experiments: evaluating xVerify as a judge model on both in-domain and out-of-distribution datasets, and using xVerify as a reward model in the reinforcement learning optimization process. First, we will outline the experimental setup:

\paragraph{Datasets:} For the evaluation experiments, we primarily use the test set and generalization set from the VAR dataset. The test set evaluates xVerify’s core performance, while the generalization set assesses its robustness on out-of-distribution samples. For the reinforcement learning experiments, the dataset for training and testing the policy model is collected from diverse sources. Further details are provided in Appendix~\ref{tag:rl_training_test_dataset}.

\paragraph{Metrics:} Accuracy and F1 are used as the main metrics to provide a comprehensive assessment.

\paragraph{Baselines:} There are two types of baselines: evaluation frameworks and judge models. The evaluation frameworks include DeepSeek-Math~\citep{deepseekmath_2024_arXiv_deepseek}, LM Eval Harness~\citep{lm_eval_harness_21_Zenodo}, Math-Verify~\citep{math_verify_24_github}, OpenAI Evals~\citep{evals_24_github_OPENAI}, OpenCompass~\citep{opencompass_24_github}, and UltraEval~\citep{ultraeval_24_Tsinghua}. The judge models include PandaLM~\citep{pandalm_24_icrl_Peking}, Auto-J~\citep{auto_j_24_iclr_JiaoTong}, Prometheus 2~\citep{prometheus_2_24_acl_KAIST}, JudgeLM~\citep{judgelm_25_icrl}, and CompassJudger~\citep{CompassJudger_1_24_arXiv_Shanghai_AI}. 
GPT-4o is also used as a judge model, with and without CoT. All prompts employed for both the judge model and xVerify can be found in Appendix~\ref{appendix:prompts for judge model} and \ref{appendix:prompt for xverify}.

\input{tables/generalization_results}
\input{tables/rl_results}

\subsection{Evaluation with xVerify as Judge Model}
We evaluated all frameworks and models on VAR’s test and generalization sets (Tables~\ref{tab:test results}–\ref{tab:general results}), where xVerify consistently achieved the best results.

\paragraph{Test Set Evaluation Results.} On the VAR test set, \textbf{xVerify consistently outperforms all evaluation frameworks and judge models}. Even the smallest xVerify-0.5B-I achieves second-best overall accuracy (96.85\%) and F1 (96.69\%), surpassing CompassJudger-1-32B and matching GPT-4o’s performance while using far fewer tokens. Larger xVerify variants (3B–32B) further improve both F1 and accuracy, peaking at 97.50\%/97.41\% (F1/Acc.) with xVerify-7B-I. Notably, all xVerify models above 0.5B exceed 95\% on challenging math questions, and performance gains taper beyond 7B parameters—suggesting a sweet spot around mid-scale models for this dataset.

\paragraph{Generalization Set Evaluation Results.} On the VAR generalization set, xVerify’s overall F1 and accuracy drop by less than 1.5\%, demonstrating strong robustness to out-of-distribution samples. Even xVerify-0.5B-I retains 95.53\% accuracy, outperforming all rule-based frameworks and most judge models except GPT-4o. Larger xVerify models reduce the performance gap further: xVerify-14B-Ia reaches 96.65\% accuracy with over 90\% on math questions. These results confirm that scaling xVerify enhances generalization, and that fine-tuned judge models can outperform CoT-based prompting without incurring extra token costs.

Supplementary experiments detailed in Appendix~\ref{appendix:additional results} provide further empirical support for the effectiveness and robustness of xVerify.

\subsection{Reinforcement Learning with xVerify as Reward Model}

We investigate whether a more accurate reward model improves optimization efficiency and final performance in RL fine-tuning. We build on veRL~\citep{sheng2024hybridflow} and train Qwen2.5-7B and Llama3.1-8B with GRPO~\citep{deepseekmath_2024_arXiv_deepseek}, using xVerify-7B-I as the reward and Math-Verify as a rule-based baseline. This experiment is a proof-of-concept designed to compare reward-signal quality and training dynamics rather than to maximize final scores. Training hyperparameters are listed in Appendix~\ref{appendix:rl hyperparameters}. For faithful evaluation, we manually annotate all samples (Appendix~\ref{appendix:Human Annotation}).

Table~\ref{tab:evaluation_results} shows that RL with xVerify substantially improves over direct generation (e.g., Qwen2.5-7B gains $18.4\%$ on the seven-benchmark average). Compared to the Math-Verify baseline, xVerify achieves higher final averages, 73.0\% versus 72.2\% for Qwen2.5-7B and 61.2\% versus 60.4\% for Llama3.1-8B. The modest improvement is consistent with the constrained proof-of-concept setting and the limited capacity of the policy models. Crucially, the training dynamics highlight xVerify’s advantage. The RL learning curves, which plot evaluation accuracy over training steps, show that for both Qwen2.5-7B and Llama3.1-8B, xVerify starts at a higher accuracy and converges in fewer steps than Math-Verify. The RL learning-curve plots are provided in Appendix~\ref{appendix:learning curve}. Moreover, Math-Verify induces reward hacking (e.g., consistently appending \textit{\textbackslash boxed\{\}}), whereas xVerify rewards correctness without enforcing brittle formats, leading to more effective learning.

Furthermore, We show that xVerify aligns closely with human judgments, with full results provided in Appendix~\ref{appendix:consistency test}.

\section{Conclusion}

In this paper, we propose xVerify, an efficient answer verifier for evaluating long reasoning responses generated by reasoning models on challenging objective questions. To train and evaluate xVerify, we construct the VAR dataset based on multiple LLMs and evaluation benchmarks, which collects long-form reasoning answers and is carefully annotated through multiple rounds of verification by GPT-4o and human annotators. Using VAR, we train xVerify models of different scales and compare them with existing evaluation frameworks and judge models on both test and generalization sets. Experimental results show that even the smallest xVerify-0.5B-I model outperforms all methods except GPT-4o, while larger xVerify models achieve the best overall performance, demonstrating strong effectiveness and generalization. Additionally, reinforcement learning experiments show that xVerify is effective as a reward model, effectively enhances policy performance compared to direct generation.

\section*{Limitations}

In this work, we focus on building a more accurate and efficient answer equivalence verifier, particularly for assessing the equivalence between the outputs of reasoning models and reference answers. However, the current xVerify model is not yet capable of fully replacing all evaluation methods and reward models. On the one hand, xVerify is currently tailored to objective questions with definitive answers and lacks effective optimization for subjective questions. On the other hand, xVerify cannot yet serve as a complete substitute for general-purpose reward models, as such models typically require only the input question and the output from the policy model to generate reward signals, whereas xVerify still depends on the availability of reference answers. Addressing these limitations may require future efforts to enhance xVerify's capability in scenarios without reference answers, such as those involving subjective questions. This could involve the development of new task settings and corresponding datasets for training and optimizing the xVerify model.

\bibliography{custom}

\clearpage
\appendix
\appendixpage

\startcontents[sections]
\printcontents[sections]{l}{1}{\setcounter{tocdepth}{2}}


\section*{Appendix}

\section{Datasets and Models}
\label{appendix:datasets and models}

This section will present the relevant information for all the public datasets and LLMs involved in the experiments of this paper.

In this study, we employ a total of 24 datasets, which are categorized into four primary types: multiple-choice questions (Choice), short answer questions (Short Answer), mathematical problems (Math), and classification tasks (Classification), as summarized in Table 
~\ref{tab:datasets}. To evaluate the multilingual capabilities of the xVerify model, each question type includes datasets in both Chinese and English, with one dataset featuring multilingual content. For each dataset, samples are partitioned into training and test sets following a 2:1 ratio, with the training and test sets ideally comprising 2,000 and 1,000 instances, respectively. In certain cases, the number of available samples is below 3,000, or the official test set is not publicly available, resulting in reduced dataset sizes after preprocessing.

\begin{table*}[!ht]
    \centering
    \resizebox{\linewidth}{!}{
    \begin{threeparttable}
    \begin{tabular}{@{}llllll@{}}
        \toprule
        \textbf{Dataset} & \textbf{Type} & \textbf{\#Train} & \textbf{\#Test} & \textbf{Language} & \textbf{License} \\ 
        \midrule
        CMMLU & Choice & 2000 & 1000 & Chinese & CC-BY-NC-4.0 \\
        C-Eval & Choice & 1346 & 260 & Chinese & CC-BY-NC-SA-4.0 \\
        GPQA & Choice & 794 & 398 & English & CC-BY-4.0 \\
        MMLU & Choice & 1816 & 1000 & English & MIT \\
        MMLU-Pro & Choice & 2000 & 1000 & English & MIT \\
        MMLU-Redux & Choice & 2000 & 1000 & English & CC-BY-4.0 \\
        AgNews & Classification & 2000 & 1000 & English & Unspecified \\
        Amazon & Classification & 2000 & 1000 & English & Apache-2.0 \\
        CLUEWSC & Classification & 1548 & 1000 & Chinese & Unspecified \\
        CMNLI & Classification & 2000 & 1000 & Chinese & Apache-2.0 \\ 
        AMC23 & Math & 26 & 14 & English & Unspecified \\
        AIME 2024 & Math & 20 & 10 & English & MIT \\
        CMATH & Math & 1128 & 565 & Chinese & CC-BY-4.0 \\
        GSM8K & Math & 2000 & 1000 & English & MIT \\
        LiveMathBench & Math & 190 & 93 & English \& Chinese & CC-BY-4.0 \\
        MATH & Math & 2000 & 1000 & English & MIT \\
        MGSM & Math & 1892 & 946 & Multilingual & CC-BY-SA-4.0 \\
        OlympiadBench & Math & 1787 & 892 & English \& Chinese & Apache-2.0 \\
        ARC & Short Answer & 2000 & 1000 & English & CC-BY-SA-4.0 \\
        CHID & Short Answer & 2000 & 1000 & Chinese & Apache-2.0 \\
        C-SimpleQA & Short Answer & 2000 & 1000 & Chinese & CC-BY-NC-SA-4.0 \\
        DROP & Short Answer & 2000 & 1000 & English & CC-BY-SA-4.0 \\
        FRAMES & Short Answer & 550 & 274 & English & Apache-2.0 \\
        SimpleQA & Short Answer & 2000 & 1000 & English & MIT \\
        \bottomrule
    \end{tabular}
    \end{threeparttable}
    }
    \caption{Datasets Description. The "Type" column indicates the question type in the corresponding dataset, including multiple-choice questions (Choice), short answer questions (Short Answer), math questions (Math), and classification questions (Classification).}
    \label{tab:datasets}
\end{table*}

A total of 19 large language models (LLMs) are utilized in our experiments, encompassing a diverse range of model sizes and types, with a particular emphasis on reasoning models (see Table ~\ref{tab:LLMs}). These models are subsequently used to collect LLM-generated responses and to train the xVerify model.

\begin{table*}[!ht]
    \centering
    \begin{threeparttable}
    \begin{tabular}{@{}lccll@{}}
        \toprule
        \textbf{Model} & \textbf{\#Para.} & \textbf{Type} & \textbf{Publisher} & \textbf{Date} \\ 
        \midrule
        ChatGLM3-6B & 6B & Chat & Tsinghua & 2023.10 \\
        GPT-4o & NaN & Chat & OpenAI & 2024.05 \\ 
        Gemma-2-2B-it & 2B & Instruct & Google & 2024.06 \\
        Gemma-2-9B-it & 9B & Instruct & Google & 2024.06 \\
        GLM-4-9B-Chat & 9B & Chat & Tsinghua & 2024.06 \\
        InternLM2.5-7B-Chat & 7B & Chat & ShLab & 2024.06 \\
        Qwen2-1.5B-Instruct & 1.5B & Instruct & Alibaba & 2024.06 \\
        Qwen2-7B-Instruct & 7B & Instruct & Alibaba & 2024.06 \\
        Llama-3.1-8B-Instruct & 8B & Instruct & Meta & 2024.07 \\
        Llama-3.2-1B-Instruct & 1B & Instruct & Meta & 2024.09 \\
        Llama-3.2-3B-Instruct & 3B & Instruct & Meta & 2024.09 \\
        Qwen2.5-7B-Instruct & 7B & Instruct & Alibaba & 2024.09 \\
        Qwen2.5-14B-Instruct & 14B & Instruct & Alibaba & 2024.09 \\
        Phi-4 & 14B & Chat & Microsoft & 2024.11 \\
        DeepSeek-R1-Distill-Llama-8B & 8B & Distill & DeepSeek & 2025.01 \\
        DeepSeek-R1-Distill-Qwen-1.5B & 1.5B & Distill & DeepSeek & 2025.01 \\
        DeepSeek-R1-Distill-Qwen-7B & 7B & Distill & DeepSeek & 2025.01 \\
        DeepSeek-R1-Distill-Qwen-14B & 14B & Distill & DeepSeek & 2025.01 \\
        QwQ-32B & 32B & Instruct & Alibaba & 2025.03 \\
        \bottomrule
    \end{tabular}
    \end{threeparttable}
    \caption{LLMs Description. LLMs are listed by release date. All models are chat or instruct type. "NaN" indicates that public data is unavailable.}
    \label{tab:LLMs}
\end{table*}

\section{VAR Dataset Details}

This section will present detailed information about the components of the VAR dataset, the details of human annotations, and examples from the dataset.

\subsection{Details of Training, Test, and Generalization Sets}
\label{appendix:three set details}

\subsubsection{Training Set}

The training set comprises 43,204 samples. Tables~\ref{tab:sample size for llm in training set} to~\ref{tab:sample size for question type in training set} provide the sample counts corresponding to each LLM, dataset, prompt template, and question type. Note that datasets with names containing "\_enh" refer to the augmented multiple choice question datasets.

\begin{table}[htbp!]
    \centering
    \begin{threeparttable}
    \begin{tabular}{@{}lc@{}}
    \toprule
    \textbf{Model} & \textbf{Sample Counts} \\ 
    \midrule
    ChatGLM3-6B & 2588 \\
    GPT-4o & 2691 \\
    Gemma-2-2B-it & 2657 \\
    Gemma-2-9B-it & 2600 \\
    GLM-4-9B-Chat & 2957 \\
    InternLM2.5-7B-Chat & 2935 \\
    Qwen2-1.5B-Instruct & 2700 \\
    Qwen2-7B-Instruct & 2898 \\
    LLaMA-3.1-8B-Instruct & 2852 \\
    Qwen2.5-7B-Instruct & 2854 \\
    Qwen2.5-14B-Instruct & 2801 \\
    DeepSeek-R1-Distill-Llama-8B & 3223 \\
    DeepSeek-R1-Distill-Qwen-1.5B & 3231 \\
    DeepSeek-R1-Distill-Qwen-7B & 3075 \\
    DeepSeek-R1-Distill-Qwen-14B & 3142 \\
    \bottomrule
    \end{tabular}
    \end{threeparttable}
    \caption{Number of samples from each LLM in the training set.}
    \label{tab:sample size for llm in training set}
\end{table}

\begin{table}[htbp!]
    \centering
    \begin{threeparttable}
    \begin{tabular}{@{}lc@{}}
    \toprule
    \textbf{Dataset} & \textbf{Sample Counts} \\ 
    \midrule
    CMMLU & 1557 \\
    CMMLU\_enh & 1641 \\
    GPQA & 1587 \\
    GPQA\_enh & 1668 \\
    MMLU & 1520 \\
    MMLU\_enh & 1513 \\
    MMLU-Pro & 1394 \\
    MMLU-Pro\_enh & 1442 \\
    AgNews & 1751 \\
    CLUEWSC & 5008 \\
    AMC23 & 1625 \\
    AIME 2024 & 1333 \\
    CMATH & 1893 \\
    GSM8K & 1836 \\
    MATH & 2485 \\
    MGSM & 1384 \\
    OlympiadBench\_en & 2573 \\
    OlympiadBench\_zh & 2709 \\
    CHID & 2424 \\
    C-SimpleQA & 1913 \\
    DROP & 1928 \\
    FRAMES & 2020 \\
    \bottomrule
    \end{tabular}
    \end{threeparttable}
    \caption{Number of samples from each dataset in the training set.}
    \label{tab:sample size for dataset in training set}
\end{table}

\begin{table}[htbp!]
    \centering
    \begin{threeparttable}
    \begin{tabular}{@{}lc@{}}
    \toprule
    \textbf{Prompt Template} & \textbf{Sample Counts} \\ 
    \midrule
    0-shot & 4884 \\
    0-shot-restrict & 5977 \\
    0-shot-cot & 4907 \\
    0-shot-cot-restrict & 6041 \\
    5-shot & 4774 \\
    5-shot-restrict & 5866 \\
    5-shot-cot & 4916 \\
    5-shot-cot-restrict & 5839 \\ 
    \bottomrule
    \end{tabular}
    \end{threeparttable}
    \caption{Number of samples from each prompt template in the training set.}
    \label{tab:sample size for prompt in training set}
\end{table}

\begin{table}[htbp!]
    \centering
    \begin{threeparttable}
    \begin{tabular}{@{}lc@{}}
    \toprule
    \textbf{Dataset} & \textbf{Sample Counts} \\ 
    \midrule
    Multiple Choice & 12322 \\
    Math & 15838 \\
    Short Answer & 8285 \\
    Classification & 6759 \\ 
    \bottomrule
    \end{tabular}
    \end{threeparttable}
    \caption{Number of samples from each question type in the training set.}
    \label{tab:sample size for question type in training set}
\end{table}

\subsubsection{Test Set}

The test set comprises 6,122 samples. Tables~\ref{tab:sample size for llm in test set} to~\ref{tab:sample size for question type in test set} provide the sample counts corresponding to each LLM, dataset, prompt template, and question type. Note that datasets with names containing "\_enh" refer to the augmented multiple choice question datasets.

\begin{table}[htbp!]
    \centering
    \begin{threeparttable}
    \begin{tabular}{@{}lc@{}}
    \toprule
    \textbf{Model} & \textbf{Sample Counts} \\ 
    \midrule
    ChatGLM3-6B & 378 \\
    GPT-4o & 400 \\
    Gemma-2-2B-it & 416 \\
    Gemma-2-9B-it & 369 \\
    GLM-4-9B-Chat & 367 \\
    InternLM2.5-7B-Chat & 367 \\
    Qwen2-1.5B-Instruct & 433 \\
    Qwen2-7B-Instruct & 427 \\
    LLaMA-3.1-8B-Instruct & 404 \\
    Qwen2.5-7B-Instruct & 374 \\
    Qwen2.5-14B-Instruct & 415 \\
    DeepSeek-R1-Distill-Llama-8B & 430 \\
    DeepSeek-R1-Distill-Qwen-1.5B & 451 \\
    DeepSeek-R1-Distill-Qwen-7B & 439 \\
    DeepSeek-R1-Distill-Qwen-14B & 452 \\
    \bottomrule
    \end{tabular}
    \end{threeparttable}
    \caption{Number of samples from each LLM in the test set.}
    \label{tab:sample size for llm in test set}
\end{table}

\begin{table}[htbp!]
    \centering
    \begin{threeparttable}
    \begin{tabular}{@{}lc@{}}
    \toprule
    \textbf{Dataset} & \textbf{Sample Counts} \\ 
    \midrule
    CMMLU & 216 \\
    CMMLU\_enh & 195 \\
    GPQA & 207 \\
    GPQA\_enh & 235 \\
    MMLU & 225 \\
    MMLU\_enh & 222 \\
    MMLU-Pro & 171 \\
    MMLU-Pro\_enh & 192 \\
    AgNews & 261 \\
    CLUEWSC & 710 \\
    AMC23 & 258 \\
    AIME 2024 & 186 \\
    CMATH & 263 \\
    GSM8K & 262 \\
    MATH & 362 \\
    MGSM & 205 \\
    OlympiadBench\_en & 349 \\
    OlympiadBench\_zh & 446 \\
    CHID & 347 \\
    C-SimpleQA & 270 \\
    DROP & 265 \\
    FRAMES & 275 \\
    \bottomrule
    \end{tabular}
    \end{threeparttable}
    \caption{Number of samples from each dataset in the test set.}
    \label{tab:sample size for dataset in test set}
\end{table}

\begin{table}[htbp!]
    \centering
    \begin{threeparttable}
    \begin{tabular}{@{}lc@{}}
    \toprule
    \textbf{Dataset} & \textbf{Sample Counts} \\ 
    \midrule
    Multiple Choice & 1663 \\
    Math & 2331 \\
    Short Answer & 1157 \\
    Classification & 971 \\ 
    \bottomrule
    \end{tabular}
    \end{threeparttable}
    \caption{Number of samples from each prompt template in the test set.}
    \label{tab:sample size for prompt in test set}
\end{table}

\begin{table}[htbp!]
    \centering
    \begin{threeparttable}
    \begin{tabular}{@{}lc@{}}
    \toprule
    \textbf{Prompt Template} & \textbf{Sample Counts} \\ 
    \midrule
    0-shot & 680 \\
    0-shot-restrict & 798 \\
    0-shot-cot & 642 \\
    0-shot-cot-restrict & 891 \\
    5-shot & 690 \\
    5-shot-restrict & 789 \\
    5-shot-cot & 702 \\
    5-shot-cot-restrict & 930 \\ 
    \bottomrule
    \end{tabular}
    \end{threeparttable}
    \caption{Number of samples from each question type in the test set.}
    \label{tab:sample size for question type in test set}
\end{table}

\subsubsection{Generalization Set}

The generalization set comprises 6,468 samples. Tables~\ref{tab:sample size for llm in general set} to~\ref{tab:sample size for question type in general set} provide the sample counts corresponding to each LLM, dataset, prompt template, and question type. Note that datasets with names containing "\_enh" refer to the augmented multiple choice question datasets.

\begin{table}[htbp!]
    \centering
    \begin{threeparttable}
    \begin{tabular}{@{}lc@{}}
    \toprule
    \textbf{Model} & \textbf{Sample Counts} \\ 
    \midrule
    ChatGLM3-6B & 300 \\
    GPT-4o & 305 \\
    Gemma-2-2B-it & 427 \\
    Gemma-2-9B-it & 296 \\
    GLM-4-9B-Chat & 339 \\
    InternLM2.5-7B-Chat & 341 \\
    Qwen2-1.5B-Instruct & 280 \\
    Qwen2-7B-Instruct & 346 \\
    LLaMA-3.1-8B-Instruct & 400 \\
    LLaMA-3.2-1B-Instruct & 314 \\
    LLaMA-3.2-3B-Instruct & 310 \\
    Qwen2.5-7B-Instruct & 326 \\
    Qwen2.5-14B-Instruct & 334 \\
    Phi-4 & 314 \\
    DeepSeek-R1-Distill-Llama-8B & 341 \\
    DeepSeek-R1-Distill-Qwen-1.5B & 399 \\
    DeepSeek-R1-Distill-Qwen-7B & 375 \\
    DeepSeek-R1-Distill-Qwen-14B & 434 \\
    QwQ-32B & 287 \\ 
    \bottomrule
    \end{tabular}
    \end{threeparttable}
    \caption{Number of samples from each LLM in the generalization set.}
    \label{tab:sample size for llm in general set}
\end{table}

\begin{table}[htbp!]
    \centering
    \begin{threeparttable}
    \begin{tabular}{@{}lc@{}}
    \toprule
    \textbf{Dataset} & \textbf{Sample Counts} \\ 
    \midrule
    C-Eval & 435 \\
    C-Eval\_enh & 442 \\
    MMLU-Redux & 436 \\
    MMLU-Redux\_enh & 483 \\
    Amazon & 646 \\
    CMNLI & 643 \\
    LiveMathBench\_en & 1127 \\
    LiveMathBench\_zh & 821 \\
    ARC & 807 \\
    SimpleQA & 628 \\ 
    \bottomrule
    \end{tabular}
    \end{threeparttable}
    \caption{Number of samples from each dataset in the generalization set.}
    \label{tab:sample size for dataset in general set}
\end{table}

\begin{table}[htbp!]
    \centering
    \begin{threeparttable}
    \begin{tabular}{@{}lc@{}}
    \toprule
    \textbf{Dataset} & \textbf{Sample Counts} \\ 
    \midrule
    Multiple Choice & 1796 \\
    Math & 1948 \\
    Short Answer & 1435 \\
    Classification & 1289 \\ 
    \bottomrule
    \end{tabular}
    \end{threeparttable}
    \caption{Number of samples from each prompt template in the generalization set.}
    \label{tab:sample size for prompt in general set}
\end{table}

\begin{table}[htbp!]
    \centering
    \begin{threeparttable}
    \begin{tabular}{@{}lc@{}}
    \toprule
    \textbf{Prompt Template} & \textbf{Sample Counts} \\ 
    \midrule
    0-shot & 703 \\
    0-shot-restrict & 856 \\
    0-shot-cot & 772 \\
    0-shot-cot-restrict & 915 \\
    5-shot & 690 \\
    5-shot-restrict & 885 \\
    5-shot-cot & 756 \\
    5-shot-cot-restrict & 891 \\ 
    \bottomrule
    \end{tabular}
    \end{threeparttable}
    \caption{Number of samples from each question type in the generalization set.}
    \label{tab:sample size for question type in general set}
\end{table}

\subsection{Details of Human Annotation}
\label{appendix:Human Annotation}

To ensure high-quality annotation for the VAR dataset, we assembled a team of 8 annotators. Among them, 6 hold bachelor's degrees and are primarily responsible for batch annotation tasks, while the other 2 hold master's degrees and focus on reviewing complex cases or resolving discrepancies in annotations made by multiple annotators. The gender ratio within the annotation team is balanced at 1:1. In terms of compensation, all annotators were paid according to the local industry average rates. The annotation process lasted for three weeks, covering a total of 15 working days.

\begin{figure*}[htbp!]
    \centering
    \includegraphics[width=1.0\linewidth]{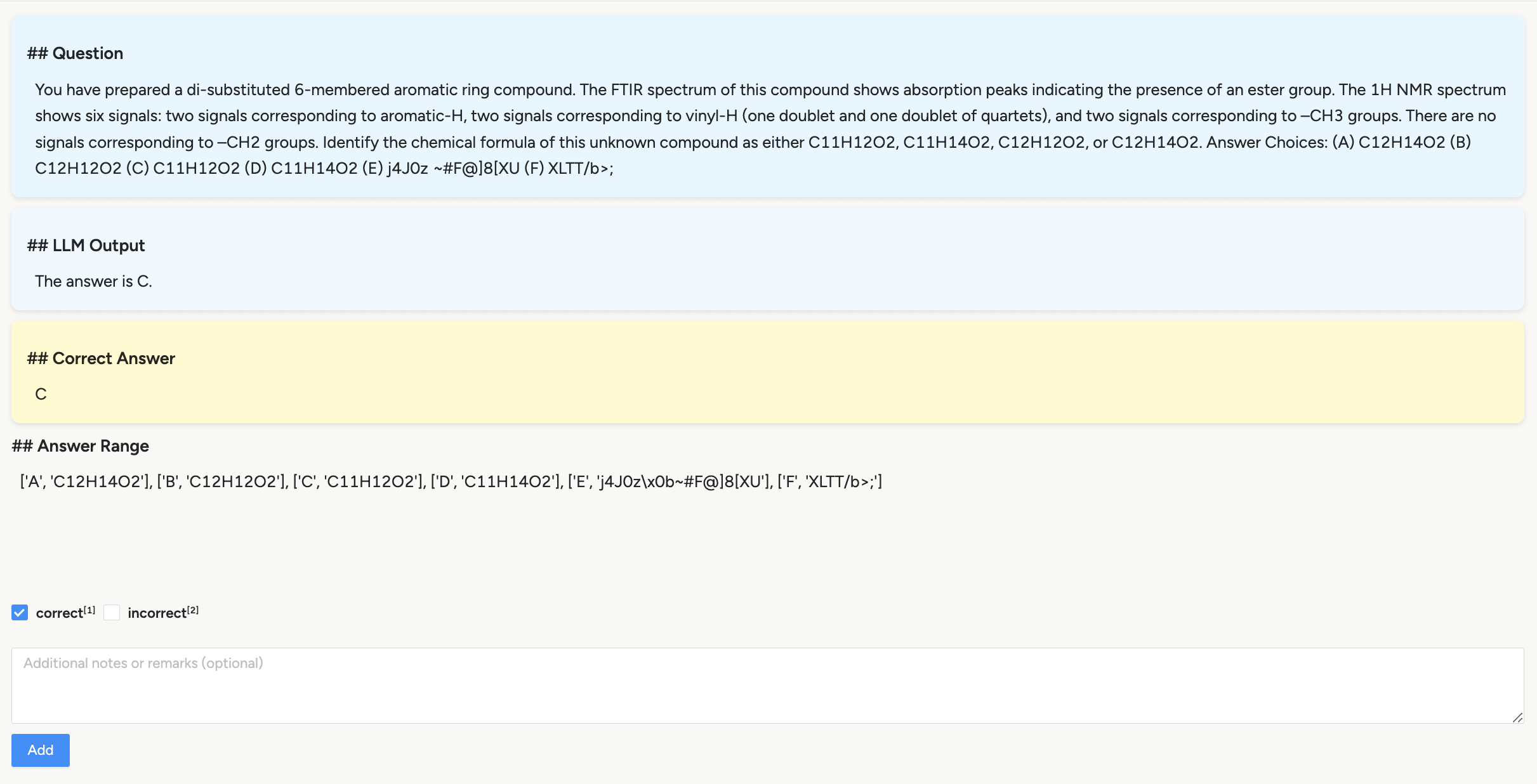}
    \caption{Illustration of the Label Studio Interface.} 
    \label{fig:appendix label-studio-interface}
\end{figure*}

The detailed annotation guidelines are presented below. Figure~\ref{fig:appendix label-studio-interface} shows an example of the interface used in our annotation tool. Each sample to be annotated contains four fields: \texttt{question}, \texttt{LLM output}, \texttt{correct answer}, and \texttt{answer range}. The \texttt{question type} includes four categories: \textit{multiple choice}, \textit{math}, \textit{short answer}, and \textit{classification}. Annotators are required to judge whether the \texttt{LLM output} matches the \texttt{correct answer} based on the \texttt{question}, while the \texttt{answer range} serves as auxiliary reference information to support the decision-making process. The specific annotation instructions and criteria are as follows:

\textbf{Answer evaluation criteria for different question types: }

\begin{itemize}
  \item \textbf{Multiple Choice} \\
  For multiple-choice questions, answer options may be labeled with letters (A, B, C, D, \ldots), Roman numerals (I, II, III, IV, \ldots), or Arabic numerals (1, 2, 3, 4, \ldots). The LLM output is considered \textit{correct} if it provides:
  \begin{itemize}
    \item Only the correct option label;
    \item Only the correct option content;
    \item Both the correct label and content.
  \end{itemize}
  In cases where the label and content are inconsistent, the \textbf{content takes precedence}. If the content is correct, the answer is marked as \textit{correct}; if the content is incorrect, the answer is marked as \textit{incorrect}, even if the option label is correct (see the final annotation example for reference).

  \item \textbf{Short Answer} \\
  Short-answer questions may require responses such as names, locations, numbers, dates, or full sentences. The evaluation criteria are:
  \begin{itemize}
    \item For concise answers (e.g., names, places, dates), strict string matching is required.
    \item For sentence-level answers, semantic consistency with the reference answer is required.
    \item For numerical answers, mathematical equivalence must be verified (e.g., ``12000'' and ``12{,}000'' are considered equivalent).
  \end{itemize}

  \item \textbf{Classification} \\
  Classification questions come with a fixed set of candidate answers. The LLM output must explicitly and exactly match the correct answer in this set to be judged as \textit{correct}.

  \item \textbf{Math} \\
  For mathematical questions, the final answer in the LLM output must be mathematically equivalent to the reference answer. Evaluation criteria include:
  \begin{itemize}
    \item If an initial answer (\texttt{ans1}) is given but followed by a derived final answer (\texttt{ans2}) through calculation, \texttt{ans2} should be used for evaluation.
    \item If the LLM output or ground-truth answer is provided in LaTeX format and cannot be visually interpreted, a LaTeX compiler should be used to determine equivalence.
  \end{itemize}
\end{itemize}

\textbf{Special cases: }

\begin{itemize}
  \item \textbf{Overly Long Responses} \\
  If the LLM output is excessively long, use the final answer provided as the basis for judgment. If the response does not converge to a clear answer (e.g., repeated changes or ambiguity), it should be marked as \textit{incorrect}.

  \item \textbf{Truncated Calculations} \\
  In long responses where the final verification or calculation is truncated, it can be ignored. If a clear answer was provided earlier, use it for evaluation; if not, mark the response as \textit{incorrect} (see the second-to-last annotation example).

  \item \textbf{Unfamiliar Domains} \\
  If the correctness of the LLM response cannot be determined due to unfamiliar domain knowledge (e.g., mathematical expressions whose equivalence is hard to judge), the case may be skipped and will later be annotated by more qualified annotators.

  \item \textbf{Irrelevant but Matched Answers} \\
  If the LLM output is irrelevant to the question but coincidentally contains the correct final answer, it should be marked as \textit{incorrect}.
\end{itemize}

\subsection{Examples from the VAR Dataset}
\label{appendix:Examples from the VAR Dataset}

In Figure~\ref{fig:appendix examples of var}, we provide several representative examples from the VAR dataset. These samples illustrate the structure of the data instances.

\lstset{
    keywordstyle=\color{blue},
    commentstyle=\color{gray},  
    stringstyle=\color{green},
    basicstyle=\small\ttfamily,
    frame=single,           
    breaklines=true,
    backgroundcolor=\color{gray!10},
    rulecolor=\color{black},
    morekeywords={
        question,
        question_type,
        llm_output,
        correct_answer,
        dataset,
        human_judgment_result
    },
    framesep=5pt,
    frameround=tttt,
    breakindent=0pt,
    escapeinside={(*@}{@*)},
}

\begin{figure*}[t]
\centering
\begin{lstlisting}
{
    "dataset": "SimpleQA",
    "question": "In which year did Fayaz A. Malik (an Indian pharmacologist, cancer biologist, and scientist) receive the Young Scientist of the Year from the Council of Scientific and Industrial Research?",
    "question_type": "short answer",
    "correct_answer": "2009",
    "llm_output": "The year Fayaz A. Malik received the Young Scientist of the Year award from the Council of Scientific and Industrial Research was 2001.\textbackslash n\textbackslash nThe answer is 2001.",
    "human_judgment_result": "Incorrect"
},
{
    "dataset": "MMLU-Redux_enh",
    "question": "If you wanted to find the global distribution of coal, you would use a  Answer Choices:  (I) reference map.  (II) topographic map.  (III) thematic map.  (IV) location map.",
    "question_type": "multiple choice",
    "correct_answer": "III",
    "llm_output": "The answer is (III) thematic map.",
    "human_judgment_result": "Correct"
},
{
    "dataset": "Amazon",
    "question": "Lightsaber Action!: This game is absolutly the best game I have ever seen! there is a little bit of violence, but most of the time you kill droids, not humans. My favorite weopon is the lightsaber, because you can deflect laser blasts.\textbackslash n Please identify the sentiment polarity of the sentence: positive or negative",
    "question_type": "classification",
    "correct_answer": "positive",
    "llm_output": "The answer is positive.",
    "human_judgment_result": "Correct"
},
{
    "dataset": "GSM8K",
    "question": "Daisy and Rose were enjoying their backyard pool with their dogs.  If there are 24 legs/paws in the pool, how many dogs do Daisy and Rose have?",
    "question_type": "math",
    "correct_answer": "5",
    "llm_output": "To find the number of dogs Daisy and Rose have, we need to first determine the total number of people and then subtract that from the total number of legs/paws.\n\nSince each person has 2 legs, we can divide the total number of legs/paws by 2 to find the number of people:\n24 legs/paws / 2 legs/person = 12 people\n\nSince Daisy and Rose are the only people in the pool, we can subtract them from the total number of people to find the number of dogs:\n12 people - 2 people = 10 dogs\n\nTherefore, Daisy and Rose have 10 dogs.",
    "human_judgment_result": "Incorrect"
}
\end{lstlisting}
\caption{Examples from the VAR Dataset}
\label{fig:appendix examples of var}
\end{figure*}

\section{Model Training Details}
\label{appendix:model training details}
This section will further present additional information about the training of the xVerify model.

\subsection{Training Hyperparameters} 
\label{appendix:train hyperparameters}

The xVerify model is trained using the QLoRA method, with consistent hyperparameter settings across all base models. The training is carried out on multiple GPU servers. Table~\ref{tab:hyperparameters} presents the key training hyperparameters.

\begin{table}[htbp!]
\centering
\begin{tabular}{cc}
\toprule
    \textbf{Hyperparameter}           & \textbf{Setting} \\ \midrule
    Per Device Train Batch Size           & 1                \\
    Gradient Accumulation Steps         & 8                \\
    Learning Rate                      & 1.0e-4           \\
    Num Train Epochs                    & 1.0              \\
    LrScheduler Type                   & cosine           \\
    Warmup Ratio                       & 0.1              \\
    Bf16                              & true             \\
    Ddp Timeout                        & 180000000        \\
    Lora Rank                          & 8                \\
\bottomrule
\end{tabular}
\caption{Hyperparameter settings for model training.}
\label{tab:hyperparameters}
\end{table}

\subsection{Original Model Details}
\label{appendix:original model details}

This paper uses 14 original models of different parameter scales and types for training on the VAR dataset. Table~\ref{tab:Details of Original Models} presents the relevant information for all xVerify models and their corresponding original models.

\begin{table*}[htbp!]
    \centering
    \begin{tabular}{@{}lcccl@{}}
    \toprule
    \textbf{Original Model} & \textbf{\#Para.} & \textbf{Type} & \textbf{Context Length} & \textbf{xVerify Model} \\ 
    \midrule
    Gemma-2-2B-it & 2B & Instruct & 8K & xVerify-2B-I \\
    Gemma-2-9B-it & 9B & Instruct & 8K & xVerify-9B-I \\
    Gemma-2-27B-it & 27B & Instruct & 8K & xVerify-27B-I \\
    GLM-4-9B-Chat & 9B & Chat & 128K & xVerify-9B-C \\
    Llama-3.2-1B-Instruct & 1B & Instruct & 128K & xVerify-1B-I \\
    Llama-3.2-3B-Instruct & 3B & Instruct & 128K & xVerify-3B-Ia \\
    Llama-3.1-8B-Instruct & 8B & Instruct & 128K & xVerify-8B-I \\
    Phi-4 & 14B & Instruct & 16k & xVerify-14B-Ib \\
    Qwen2.5-0.5B-Instruct & 0.5B & Instruct & 128K & xVerify-0.5B-I \\
    Qwen2.5-1.5B-Instruct & 1.5B & Instruct & 128K & xVerify-1.5B-I \\
    Qwen2.5-3B-Instruct & 3B & Instruct & 128K & xVerify-3B-Ib \\
    Qwen2.5-7B-Instruct & 7B & Instruct & 128K & xVerify-7B-I \\
    Qwen2.5-14B-Instruct & 14B & Instruct & 128K & xVerify-14B-Ia \\
    Qwen2.5-32B-Instruct & 32B & Instruct & 128K & xVerify-32B-I \\ 
    \bottomrule
    \end{tabular}
    \caption{Details of Original Models and Corresponding xVerify Models. Sorted by Original Model Name.}
    \label{tab:Details of Original Models}
\end{table*}

\section{RL Training Details}
\label{appendix:rl training details}
This section will further present additional information about RL training.
\subsection{RL Training Hyperparameters} 
\label{appendix:rl hyperparameters}

We train our models using the veRL~\citep{sheng2024hybridflow} Framework. The hyperparameters for implementing RL are presented in Table \ref{tab:rl_hyperparameters}. When implementing baseline methods, we use the same hyperparameters. The training is carried out on multiple GPU servers. 

\begin{table}[htbp!]
\centering
\begin{tabular}{cc}
\toprule
    \textbf{Hyperparameter}           & \textbf{Setting} \\ \midrule
    Epoch                     & 15                \\
    Learning Rate             & 1.0e-6           \\
    Train Batch Size           & 1024           \\
    Maximum Rollout Length    & 8192      \\
     KL Loss Coefficient       & 1e-3      \\
     Entropy Coefficient       & 0.0     \\
    Temperature              & 1.0     \\
    Rollouts per Prompt     & 5 \\

\bottomrule
\end{tabular}
\caption{Hyperparameter settings for RL training.}
\label{tab:rl_hyperparameters}
\end{table}

\subsection{Details of RL Training and Generalization Sets}
\label{tag:rl_training_test_dataset}
When training policy, we use xVerify as the reward model. To ensure that the evaluation truly reflects the generalization capability of xVerify, we select all training data from the generalization set, thereby avoiding data contamination. The training set comprises 20,400 samples. The specific composition is shown in Table~\ref{tab:sample size for dataset in rl training set}. The generalization set comprises 1,750 samples as shown in Table~\ref{tab:sample size for dataset in rl test set}. 

\begin{table}[htbp!]
    \centering
    \begin{threeparttable}
    \begin{tabular}{@{}lc@{}}
    \toprule
    \textbf{Dataset} & \textbf{Sample Counts} \\ 
    \midrule
MMLU-Pro & 3400 \\
MATH & 3400 \\
DROP & 3400 \\
GSM8K & 3400 \\
AgNews & 3400 \\
ARC & 3400 \\
    \bottomrule
    \end{tabular}
    \end{threeparttable}
    \caption{Number of samples from each dataset in RL training set.}
    \label{tab:sample size for dataset in rl training set}
\end{table}
 
\begin{table}[htbp!]
    \centering
    \begin{threeparttable}
    \begin{tabular}{@{}lc@{}}
    \toprule
    \textbf{Dataset} & \textbf{Sample Counts} \\ 
    \midrule
    MATH-500 & 470 \\
    MMLU-Pro & 250 \\
    DROP & 250 \\
    GPQA & 250 \\
    Amazon & 250 \\
    CMMLU & 250 \\
    CHID & 250 \\
    \bottomrule
    \end{tabular}
    \end{threeparttable}
    \caption{Number of samples from each dataset in RL generalization set.}
    \label{tab:sample size for dataset in rl test set}
\end{table}

\clearpage

\section{Prompts}
\label{appendix:prompts}

This section will present all the prompt templates used in the experiments of this paper.

\subsection{Prompts for Generating LLM Responses}
\label{appendix:prompts for generating LLM responses}

The prompt templates used to generate LLM responses are illustrated in Figures~\ref{img:prompt_few_shot} to~\ref{img:prompt_few_shot_cot_restrict}. Each template consists of four fields that need to be populated: "task\_type", "task\_description", "examples", and "question". The "task\_type" and "task\_description" fields are determined based on the type of question. For instance, for questions from the GPQA dataset, "task\_type" is set to "multidisciplinary question", and "task\_description" is set to "Please choose the answer from options A to D, corresponding to the question." During dataset preprocessing, we design appropriate "task\_type" and "task\_description" values for each dataset. The "examples" field is filled according to the selected prompting strategy, either 0-shot or 5-shot. In the 0-shot setting, this field is left empty, while in the 5-shot setting, it is populated with five example question-answer pairs that are similar to the target "question". The "question" field contains the specific query to be answered by the LLM. Examples of the "examples" and "question" fields are shown in Figures~\ref{img:example of examples fields} and~\ref{img:example of question fields}, respectively.

\begin{figure}[htbp!]
    \small
    \centering
    \begin{tcolorbox}
    You are an expert in \{task\_type\}, \{task\_description\} \\
    \{examples\} \\
    \{question\}
    
    \end{tcolorbox}
    \caption{Few-shot prompt for generating LLM responses.}
    \label{img:prompt_few_shot}
\end{figure}

\begin{figure}[htbp!]
    \small
    \centering
    \begin{tcolorbox}
    You are an expert in \{task\_type\}, \{task\_description\} \\
    \{examples\} \\
    \{question\} \\
    \\
    End your final answer with 'The answer is <answer>.'
    
    \end{tcolorbox}
    \caption{Few-shot-restrict prompt for generating LLM responses.}
    \label{img:prompt_few_shot_restrict}
\end{figure}

\begin{figure}[htbp!]
    \small
    \centering
    \begin{tcolorbox}
    You are an expert in \{task\_type\}, \{task\_description\} \\
    \{examples\} \\
    \{question\} \\
    \\
    Let’s think step by step.
    
    \end{tcolorbox}
    \caption{Few-shot-cot prompt for generating LLM responses.}
    \label{img:prompt_few_shot_cot}
\end{figure}

\begin{figure}[htbp!]
    \small
    \centering
    \begin{tcolorbox}
    You are an expert in \{task\_type\}, \{task\_description\} \\
    \{examples\} \\
    \{question\} \\
    \\
    Let’s think step by step. \\
    \\
    End your final answer with 'The answer is <answer>.'
    \end{tcolorbox}
    \caption{Few-shot-cot-restrict prompt for generating LLM responses.}
    \label{img:prompt_few_shot_cot_restrict}
\end{figure}

\begin{figure*}[!htbp]
    \small
    \centering
    \begin{tcolorbox}
    ***** Start In-Context Examples ***** \\
    Q: A late game rally by Washington led them to the Eagles' 26 yard line. A shot to the end zone by Robert Griffin III would be intercepted by Brandon Boykin, clinching an Eagles win. The Eagles would move to 6-5. This is the Eagles first win at Lincoln Financial Field since Week 4 of the 2012 season, because prior to this game, the Eagles had never won a game in their home stadium in 414 days since that same week, snapping a 10-game losing streak at home with this win. How many more wins than losses did the Eagles have after this game? \\
    A: The answer is 1. \\
    \\
    Q: The population of Sevastopol proper is 418,987 (01.01.16), making it the largest in the Crimean Peninsula. The citys agglomeration has about 600,000 people (2015). According to the Ukrainian Census (2001), the ethnic groups of Sevastopol include Russians (71.6\%), Ukrainians (22.4\%), Belarusians (1.6\%), Tatars (0.7\%), Crimean Tatars (0.5\%), Armenians (0.3\%), Jews (0.3\%), Moldovans (0.2\%), and Azerbaijani people (0.2\%). Which ethnic has a higher percentage of the population in Sevastopol: Russians or Armenians? \\
    A: The answer is Russians. \\
    \\
    Q: the most common crimes in the ACT are property related crimes, unlawful entry with intent and motor vehicle theft. They affected 2,304 and 966 people (580 and 243 per 100,000 persons respectively). Homicide and related offences—murder, attempted murder and manslaughter, but excluding driving causing death and conspiracy to murder—affect 1.0 per 100,000 persons, which is below the national average of 1.9 per 100,000. Rates of sexual assault (64.4 per 100,000 persons) are also below the national average (98.5 per 100,000). Which was there a higher national average for, homicide and related offences or sexual assault? \\
    A: The answer is sexual assault. \\
    \\
    Q: In the county, the population was spread out with 21.7\% under the age of 18,  8.5\% from 18 to 24, 26.9\% from 25 to 44, 27.7\% from 45 to 64, and 15.0\% who were 65 years of age or older. The median age was 40 years. For every 100 females, there were 94.4 males. For every 100 females age 18 and over, there were 98.7 males. How many percent were not from 45 to 64? \\
    A: The answer is 72.3. \\
    \\
    Q: The median age in the city was 35.1 years. 24.2\% of residents were under the age of 18; 7.9\% were between the ages of 18 and 24; 33.8\% were from 25 to 44; 24.6\% were from 45 to 64; and 9.5\% were 65 years of age or older. The gender makeup of the city was 48.6\% male and 51.4\% females. How many more people, in terms of percentage, were in the largest age group compared to the second smallest? \\
    A: The answer is 24.3. \\
    ***** End In-Context Examples *****
    \end{tcolorbox}
    \caption{Example of "examples" fields.}
    \label{img:example of examples fields}
\end{figure*}

\begin{figure*}[!htbp]
    \small
    \centering
    \begin{tcolorbox}
    Q: Let $ABCD$ be a tetrahedron such that $AB=CD= \sqrt{41}$, $AC=BD= \sqrt{80}$, and $BC=AD= \sqrt{89}$. There exists a point $I$ inside the tetrahedron such that the distances from $I$ to each of the faces of the tetrahedron are all equal. This distance can be written in the form $\frac{m \sqrt n}{p}$, where $m$, $n$, and $p$ are positive integers, $m$ and $p$ are relatively prime, and $n$ is not divisible by the square of any prime. Find $m+n+p$. \\
    A: 
    \end{tcolorbox}
    \caption{Example of "question" fields.}
    \label{img:example of question fields}
\end{figure*}

\subsection{Prompts for GPT-4o Annotation}
\label{appendix:prompts for GPT-4o annotation}

The prompt templates used for annotating the collected LLM question-answer pairs with GPT-4o during the construction of the VAR dataset are shown in Figures~\ref{img:prompt I for GPT-4o annotation} and~\ref{img:prompt II for GPT-4o annotation}. Both of these prompt templates employ the Chain-of-Thought (CoT) strategy to ensure the accuracy of the annotations generated by GPT-4o.

\begin{figure*}[!htbp]
    \small
    \centering
    \begin{tcolorbox}
    \textcolor{red}{
    You are a diligent and precise assistant tasked with evaluating the correctness of responses. Think step by step as you make your evaluation. \\
    }
    \\
    You will receive a question, an output sentence, and the correct answer. Your task is to determine if the output sentence accurately answers the question based on the provided correct answer. Think step by step and respond with either [Correct] or [Incorrect]. \\
    - \\
    Special considerations: \\
    1. **Multiple Answers**: If the output contains multiple answers, evaluate whether later answers modify or correct earlier ones. In such cases, compare the final answer with the correct answer. If the final answer is unclear or incorrect, respond with [Incorrect]. \\
    2. **Mathematical Problems**: If the formats differ but the answers are mathematically equivalent, respond with [Correct]. \\
    3. **Explicit Options**: If the question provides explicit candidate answers, the output will be considered correct if it clearly indicates the correct option's code or the correct option's content. \\
    4. **No Explicit Options**: If the question does not provide explicit options, the output must align with the correct answer in content and meaning to be considered [Correct]. \\
    Please present your response in the following JSON format:  \\
    \{ \\
    \verb|    |"reasoning": "Your step-by-step reasoning here.",  \\
    \verb|    |"judgment": "Correct or Incorrect" \\
    \} \\
    - \\
    Question: """\{question\}""" \\
    Output sentence: """\{output\}""" \\
    Correct answer: \{answer\} \\
    \end{tcolorbox}
    \caption{Prompt I for GPT-4o annotation.}
    \label{img:prompt I for GPT-4o annotation}
\end{figure*}

\begin{figure*}[!htbp]
    \small
    \centering
    \begin{tcolorbox}
    \textcolor{red}{
    You are a diligent and precise assistant tasked with evaluating the correctness of responses. Think step by step as you make your evaluation. \\
    }
    \\
    We request your feedback on whether the model's response correctly answers the user question above. Follow these steps to make your evaluation: \\
    1. Think step by step: Read the user question carefully. \\
    2. Think step by step: Review the reference answer and understand the key points it covers. \\
    3. Think step by step: Compare the model's answer with the reference answer. \\
    4. Think step by step: Determine if the model's answer addresses the key points in the reference answer and correctly answers the question. \\
    - \\
    First, provide your reasoning in detail. Then, clearly state your judgment as either "Correct" or "Incorrect." \\
    Please present your response in the following JSON format:  \\
    \{ \\
    \verb|    |"reasoning": "Your step-by-step reasoning here.",  \\
    \verb|    |"judgment": "Correct or Incorrect" \\
    \} \\
    - \\
    Question: \{question\} \\
    Reference Answer: \{answer\} \\
    Model's Answer: \{output\}
    \end{tcolorbox}
    \caption{Prompt II for GPT-4o annotation.}
    \label{img:prompt II for GPT-4o annotation}
\end{figure*}

\subsection{Prompts for Data Augmentation}
\label{appendix:prompts for data augmentation}

In constructing the VAR dataset, two prompt templates used to guide GPT-4o in augmenting mathematical question samples are presented in Figures~\ref{img:prompt for gene ref answer} and~\ref{img:prompt for gene final answer}.

\begin{figure*}[!htbp]
    \small
    \centering
    \begin{tcolorbox}
    \begin{verbatim}
You are an expert in mathematical calculations and data expressions.
You are required to provide different equivalent forms of the standard
answer for the following math problem.
Problem: {question}
Answer: {answer}

Example 1:
Problem: Let $\alpha$ be the radian measure of the smallest angle in a 
$3-4-5$ right triangle. Let $\beta$ be the radian measure of the 
smallest angle in a $7-24-25$ right triangle. Express $\beta$ in terms 
of $\alpha$.
Answer: $\frac{\pi}{2} - 2\alpha$
Output:
```json {
    "answer1": "\pi/2 - 2\alpha",
    "answer2": "pi/2 - 2alpha",
    "answer3": "pi/2 - 2 * alpha",
    "answer4": "0.5 * pi - 2 * alpha" }```

Example 2:
Problem: A volcano erupts and spews ash into the sky. The ash cloud 
spreads out in a diameter eighteen times as far as the distance it 
shot up into the sky. If the ashes erupted three hundred feet into the
sky, what was the radius of the ash cloud in feet?
Answer: 2700
Output:
```json {
    "answer1": "2.7×10^3",
    "answer2": "2700.0",
    "answer3": "2.7 \times 10^3",
    "answer4": "$2.7 \times 10^3$",
    "answer5": "Two thousand seven hundred" }```

Please note:
1. You need to provide 3 to 5 different standard forms of the answer.
2. Each different form must be equivalent to the standard answer, i.e.,
it should still be a correct and valid answer.
3. You may use LaTeX, scientific notation, or other standard
mathematical expressions.
4. Please follow the JSON format below for the output:
```json {
    "answer1": "xxx", "answer2": "xxx", "answer3": "xxx", ...
}```
    \end{verbatim}
    \end{tcolorbox}
    \caption{Prompt for Generating Alternative Reference Answers.}
    \label{img:prompt for gene ref answer}
\end{figure*}

\begin{figure*}[!htbp]
    \small
    \centering
    \begin{tcolorbox}
    \begin{verbatim}
You are an expert in mathematical calculations and data expressions. 
For an answer to a specific mathematical problem, you are required to 
provide equivalent and different expressions of the mathematical 
result.
Answer: {output}

Example 1:
Answer: The answer is $\beta = \frac{\pi}{2} - 2\alpha$.
Output:
```json {
    "answer1": "The answer is \pi/2 - 2\alpha.",
    "answer2": "The answer is pi/2 - 2alpha.",
    "answer3": "The answer is pi/2 - 2 * alpha.",
    "answer4": "The answer is 0.5 * pi - 2 * alpha."
}```

Example 2:
Answer: The answer is 2700 feet.
Output:
```json {
    "answer1": "The answer is 2.7×10^3 feet.",
    "answer2": "The answer is 2700.0 feet.",
    "answer3": "The answer is 2.7 \times 10^3 feet.",
    "answer4": "The answer is $2.7 \times 10^{3}$ feet.",
    "answer5": "The answer is Two thousand seven hundred feet."
}```

Please note:
1. You need to provide 3 to 5 different expressions, each replacing 
the mathematical result with an equivalent and different form.
2. Each expression must be exactly equivalent to the target answer to
ensure its correctness.
3. You can use LaTeX, scientific notation, or other standard 
mathematical formats.
4. Please output the result in the following JSON format:
```json {
    "answer1": "The answer is xxx",
    "answer2": "The answer is xxx",
    "answer3": "The answer is xxx",
    "answer4": "The answer is xxx",
    "answer5": "The answer is xxx"
}```
    \end{verbatim}
    \end{tcolorbox}
    \caption{Prompt for Generating Diverse Final Answer Expressions.}
    \label{img:prompt for gene final answer}
\end{figure*}

\subsection{Prompts for Judge Model}
\label{appendix:prompts for judge model}

In the experiments of this paper, the prompts used for all judge models were constructed based on the official templates provided by their respective developers. However, for some judge models, the official prompt templates were not fully compatible with the evaluation tasks in this paper, so other similar prompt templates were used. Specifically, Figure~\ref{img:prompt for GPT-4o as Judge} shows the prompt template used by GPT-4o as Judge, Figure~\ref{img:prompt for GPT-4o as Judge (CoT)} shows the prompt template used by GPT-4o as Judge (CoT), Figure~\ref{img:prompt for JudgeLM} shows the prompt template used by JudgeLM series models and PandaLM-7B-v1, Figure~\ref{img:prompt for Auto-J} shows the prompt template used by Auto-J series models, and Figure~\ref{img:prompt for Prometheus} shows the prompt template used by Prometheus series models. The official prompt template for the CompassJudger-1 series models corresponds to pairwise evaluation, so the prompt template used by this series is the same as that for the xVerify model, as shown in Figure~\ref{img:prompt for xverify}.

\begin{figure*}[!htbp]
    \small
    \centering
    \begin{tcolorbox}
    You are a diligent and precise assistant tasked with evaluating the correctness of responses. You will receive a question, an output sentence, and the correct answer. Your task is to determine if the output sentence accurately answers the question based on the provided correct answer. Respond with either [Correct] or [Incorrect]. \\
    - \\
    Special considerations: \\
    1. **Multiple Answers**: If the output contains multiple answers, evaluate whether later answers modify or correct earlier ones. In such cases, compare the final answer with the correct answer. If the final answer is unclear or incorrect, respond with [Incorrect]. \\
    2. **Mathematical Problems**: If the formats differ but the answers are mathematically equivalent, respond with [Correct]. \\
    3. **Explicit Options**: If the question provides explicit candidate answers, the output will be considered correct if it clearly indicates the correct option's code or the correct option's content. \\
    4. **No Explicit Options**: If the question does not provide explicit options, the output must align with the correct answer in content and meaning to be considered [Correct]. \\
    Please present your response in the following JSON format:  \\
    \{ \\
    \verb|    |"judgement": "Correct or Incorrect" \\
    \} \\
    - \\
    Question: """\{question\}""" \\
    Output sentence: """\{response\}""" \\
    Correct answer: \{reference\}
    \end{tcolorbox}
    \caption{Prompt for GPT-4o as Judge.}
    \label{img:prompt for GPT-4o as Judge}
\end{figure*}

\begin{figure*}[!htbp]
    \small
    \centering
    \begin{tcolorbox}
    You are a diligent and precise assistant tasked with evaluating the correctness of responses. Think step by step as you make your evaluation. You will receive a question, an output sentence, and the correct answer. Your task is to determine if the output sentence accurately answers the question based on the provided correct answer. Think step by step and respond with either [Correct] or [Incorrect]. \\
    - \\
    Special considerations: \\
    1. **Multiple Answers**: If the output contains multiple answers, evaluate whether later answers modify or correct earlier ones. In such cases, compare the final answer with the correct answer. If the final answer is unclear or incorrect, respond with [Incorrect]. \\
    2. **Mathematical Problems**: If the formats differ but the answers are mathematically equivalent, respond with [Correct]. \\
    3. **Explicit Options**: If the question provides explicit candidate answers, the output will be considered correct if it clearly indicates the correct option's code or the correct option's content. \\
    4. **No Explicit Options**: If the question does not provide explicit options, the output must align with the correct answer in content and meaning to be considered [Correct]. \\
    Please present your response in the following JSON format:  \\
    \{ \\
    \verb|    |"reasoning": "Your step-by-step reasoning here.", \\
    \verb|    |"judgement": "Correct or Incorrect" \\
    \} \\
    - \\
    Question: """\{question\}""" \\
    Output sentence: """\{response\}""" \\
    Correct answer: \{reference\}
    \end{tcolorbox}
    \caption{Prompt for GPT-4o as Judge (CoT).}
    \label{img:prompt for GPT-4o as Judge (CoT)}
\end{figure*}

\begin{figure*}[!htbp]
    \small
    \centering
    \begin{tcolorbox}
    You are a helpful and precise assistant for checking the quality of the answer. \\
    \verb|[Question]| \\
    \{question\} \\
    \verb|[Reference Answer]| \\
    \{reference\} \\
    \verb|[Model's Answer]| \\
    \{response\} \\
    \verb|[System]| \\
    We would like to request your feedback on the performance of the model's response to the user question displayed above. \\
    Based on the reference answer, please rate the accuracy of the response. The model receives an overall score on a scale of 1 to 10, where a higher score indicates better overall performance. \\
    Please first output a single line containing only the score. In the subsequent line, please provide a comprehensive explanation of your evaluation, avoiding any potential bias. \\
     \\
    \#\#\# Response:
    \end{tcolorbox}
    \caption{Prompt for JudgeLM.}
    \label{img:prompt for JudgeLM}
\end{figure*}

\begin{figure*}[!htbp]
    \small
    \centering
    \begin{tcolorbox}
    \verb|[INST]| Write critiques for a submitted response on a given user's query, incorporating the correct answer as a reference, and grade the response accordingly: \\
    \\
    \verb|[BEGIN DATA]| \\
    *** \\
    \verb|[Query]|: \{question\} \\
    *** \\
    \verb|[Correct Answer]|: \{reference\} \\
    *** \\
    \verb|[Response]|: \{response\} \\
    *** \\
    \verb|[END DATA]| \\
     \\
    Write critiques for this response. After that, you should give a final rating for the response on a scale of 1 to 10 by strictly following this format: "[[rating]]", for example: "Rating: [[5]]". \verb|[/INST]|
    \end{tcolorbox}
    \caption{Prompt for Auto-J.}
    \label{img:prompt for Auto-J}
\end{figure*}

\begin{figure*}[!htbp]
    \small
    \centering
    \begin{tcolorbox}
    You are a fair judge assistant tasked with providing clear, objective feedback based on specific criteria, ensuring each assessment reflects the absolute standards set for performance." \\
    \#\#\#Task Description: \\
    An instruction (might include an Input inside it), a response to evaluate, a reference answer that gets a score of 5, and a score rubric representing a evaluation criteria are given. \\
    1. Write a detailed feedback that assess the quality of the response strictly based on the given score rubric, not evaluating in general. \\
    2. After writing a feedback, write a score that is an integer between 1 and 5. You should refer to the score rubric. \\
    3. The output format should look as follows: "Feedback: (write a feedback for criteria) [RESULT] (an integer number between 1 and 5)"
    4. Please do not generate any other opening, closing, and explanations. \\
     \\
    \#\#\#The instruction to evaluate: \\
    \{question\} \\
     \\
    \#\#\#Response to evaluate: \\
    \{response\} \\
     \\
    \#\#\#Reference Answer (Score 5): \\
    \{reference\} \\
     \\
    \#\#\#Score Rubrics: \\
    \verb|[|Does the model demonstrate logical and effective reasoning in its responses?\verb|]| \\
    Score 1: The model's responses show a complete lack of logical reasoning, often resulting in irrelevant or nonsensical answers. \\
    Score 2: The model occasionally shows signs of logical reasoning but generally struggles to provide coherent or relevant responses. \\
    Score 3: The model usually demonstrates basic reasoning capabilities, though it may not consistently apply logical principles or fully resolve complex issues. \\
    Score 4: The model frequently exhibits strong reasoning skills, effectively addressing complex questions with minor inconsistencies or errors. \\
    Score 5: The model consistently demonstrates advanced reasoning abilities, providing logically sound, coherent, and sophisticated responses to complex queries. \\
     \\
    \#\#\#Feedback:  \\
    \end{tcolorbox}
    \caption{Prompt for Prometheus.}
    \label{img:prompt for Prometheus}
\end{figure*}

\subsection{Prompts for xVerify}
\label{appendix:prompt for xverify}

Figure~\ref{img:prompt for xverify} shows the prompt template used to construct the input for the xVerify model. This template is used both for training and evaluation of the xVerify model. Specifically, "question," "output," and "answer" correspond to the question content, the LLM response, and the reference answer, respectively.

\begin{figure*}[!htbp]
    \small
    \centering
    \begin{tcolorbox}
    You are a diligent and precise assistant tasked with evaluating the correctness of responses. You will receive a question, an output sentence, and the correct answer. Your task is to determine if the output sentence accurately answers the question based on the provided correct answer. Respond with either [Correct] or [Incorrect]. \\
    - \\
    Special considerations: \\
    1. **Multiple Answers**: If the output contains multiple answers, evaluate whether later answers modify or correct earlier ones. In such cases, compare the final answer with the correct answer. If the final answer is unclear or incorrect, respond with [Incorrect]. \\
    2. **Mathematical Problems**: If the formats differ but the answers are mathematically equivalent, respond with [Correct]. \\
    3. **Explicit Options**: If the question provides explicit candidate answers, the output will be considered correct if it clearly indicates the correct option's code or the correct option's content. \\
    4. **No Explicit Options**: If the question does not provide explicit options, the output must align with the correct answer in content and meaning to be considered [Correct]. \\
    - \\
    Question: """\{question\}""" \\
    Output sentence: """\{output\}""" \\
    Correct answer: \{answer\} \\
    Judgement: 
    \end{tcolorbox}
    \caption{Prompt for xVerify.}
    \label{img:prompt for xverify}
\end{figure*}


\section{Supplementary Experimental Results}
\label{appendix:additional results}

\input{tables/appendix_all_models_test_results}

\input{tables/appendix_all_models_generalization_results}

\subsection{Evaluation Accuracy Results of All xVerify Models and Judge Models}
\label{appendix:results of all models}

Tables~\ref{tab:appendix test results} and \ref{tab:appendix general results} present the performance of all xVerify models and judge models on the test set and generalization set, respectively. Overall, each xVerify model achieves an F1 score and accuracy exceeding 96.5\% on the test set, and an accuracy greater than 95.52\% on the generalization set. These results not only demonstrate the effectiveness of xVerify models in the evaluation task—consistently outperforming all other judge models—but also underscore the high quality of the VAR dataset.

A comparison of the results across the two datasets reveals that the performance of xVerify models on the generalization set exhibits a slight decline relative to the test set, with a maximum drop of no more than 1.6\%. Moreover, xVerify models with larger parameter sizes tend to show a smaller performance degradation, indicating strong generalization capabilities that further improve with model scale. Additionally, it is observed across both datasets that while the performance of xVerify models generally enhances with the increment of parameter size, beyond a certain threshold, further increases in parameter scale do not lead to additional performance gains.

\input{tables/appendix_time_costs}

\subsection{Computational Efficiency and Operational Cost of xVerify and Judge Models}
\label{appendix:efficiency and cost analyse}

Table~\ref{tab:appendix time cost results} displays the running time performance of the xVerify model and other judge models. Each model was evaluated using 200 randomly selected samples per question type from the generalization set, with running times measured in seconds. This data provides insights into the computational efficiency of each model under uniform testing conditions, thereby facilitating a comparative analysis of their real-time processing capabilities and scalability in practical applications.

All models were executed on GPUs with identical configurations. Specifically, Prometheus-8x7B-v2.0, JudgeLM-33B-v1.0, CompassJudger-1-32B, xVerify-27B-I, and xVerify-32B-I were deployed on two GPUs for inference, while the remaining models were deployed on a single GPU. From Table~\ref{tab:appendix time cost results}, it is evident that all xVerify models exhibit an overall average runtime within 100 seconds, whereas the overall average runtime for the other judge models exceeds 100 seconds. Moreover, for each question category, the models with the shortest evaluation times are the xVerify models. Thus, the xVerify models demonstrably surpass the other judge models in terms of evaluation efficiency.

Table~\ref{tab:appendix cost of gpt} presents the evaluation costs incurred when employing GPT-4o as the judge, based on assessments of 200 randomly selected samples per question type, along with the overall expenditure. Apart from the prerequisite deployment overhead, the cost of invoking the xVerify models for evaluation is substantially lower than that of GPT-4o. Additionally, compared to GPT-4o, which relies on remote server deployment, the locally deployed xVerify models offer higher invocation efficiency. Taken together, these results underscore that the xVerify models outperform the other judge models in both usage cost and evaluation efficiency.

\begin{table*}[!htbp]
    \centering
    \resizebox{.9\linewidth}{!}{
    \begin{threeparttable}
        \begin{tabular}{@{}lccccc@{}}
        \toprule
        \multicolumn{1}{c}{\textbf{Method}} & \textbf{Multiple Choice (\$)} & \textbf{Math (\$)} & \textbf{Short Answer (\$)} & \textbf{Classification (\$)} & \textbf{Total (\$)} \\ 
        \midrule
        \textbf{GPT-4o as Judge} & 0.31 & 0.66 & 0.24 & 0.27 & 1.48 \\
        \textbf{GPT-4o as Judge (CoT)} & 0.55 & 1.00 & 0.42 & 0.48 & 2.45 \\ 
        \bottomrule
        \end{tabular}
    \end{threeparttable}
    }
    \caption{Total costs (in USD) of GPT-4o as Judge (200 Samples per Question Type).}
    \label{tab:appendix cost of gpt}
\end{table*}

\subsection{The Consistency between xVerify and Human Evaluation}
\label{appendix:consistency test}
To evaluate the effectiveness of the xVerify, we conduct a comparative analysis of its consistency against human-annotated results. We evaluate the consistency between different models and human judgments across multiple evaluation datasets. The evaluation datasets and models utilized in this experiment are consistent with those employed in RL experiments (details in Appendix \ref{appendix:rl training details}). Figure~\ref{fig:consistency} illustrates that xVerify consistently achieves a high degree of alignment with human evaluation across multiple datasets and in diverse scenarios.

\begin{figure*}[!htbp]
    \centering
    \includegraphics[width=0.9\linewidth]{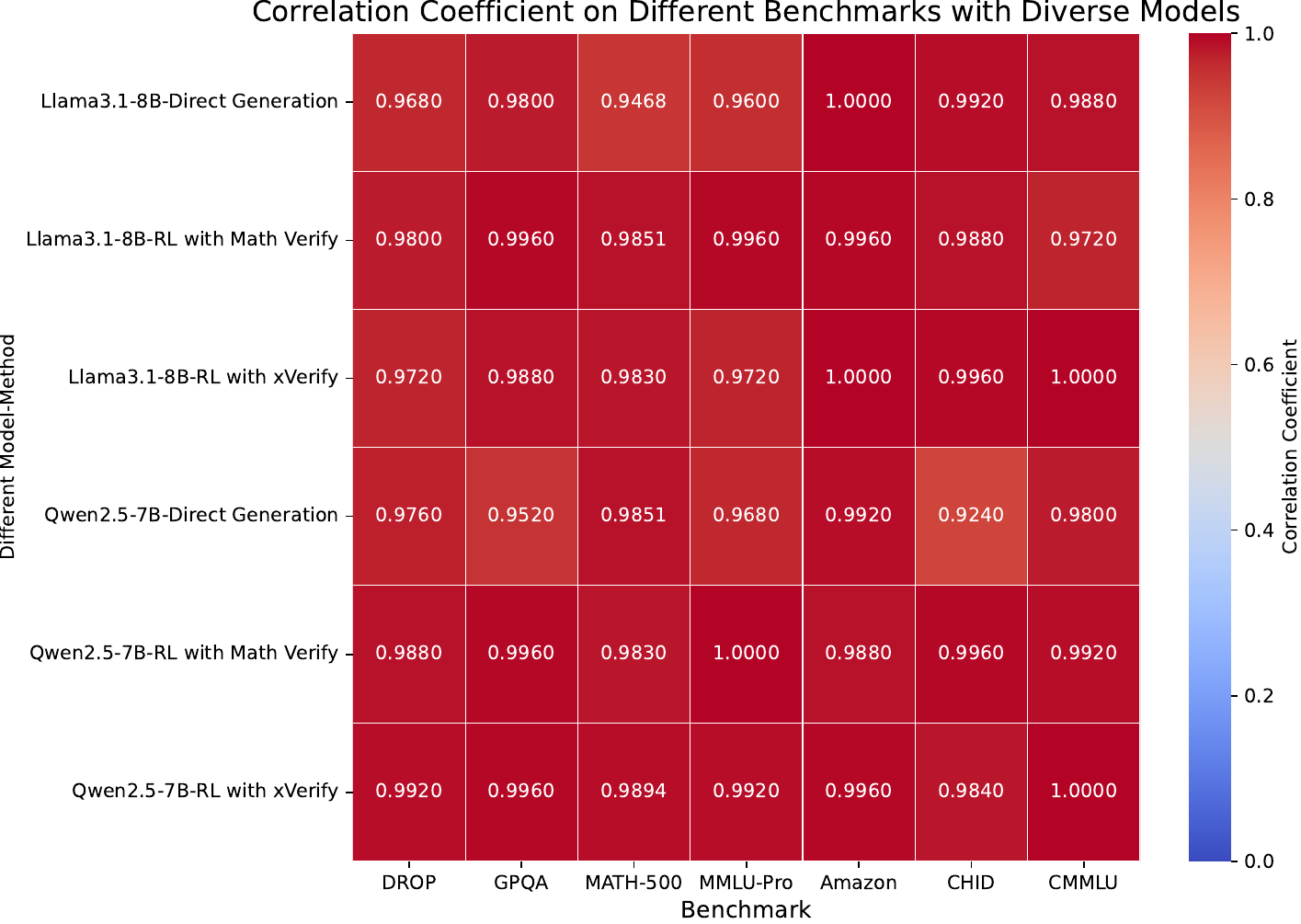}
    \caption{The Consistency between xVerify and Human Evaluation}
    \label{fig:consistency}
\end{figure*}

\subsection{Learning Curves of Qwen2.5-7B and Llama3.1-8B in Reinforcement Learning}
\label{appendix:learning curve}

\begin{figure*}[!ht]
    \centering
    \includegraphics[width=0.5\linewidth]{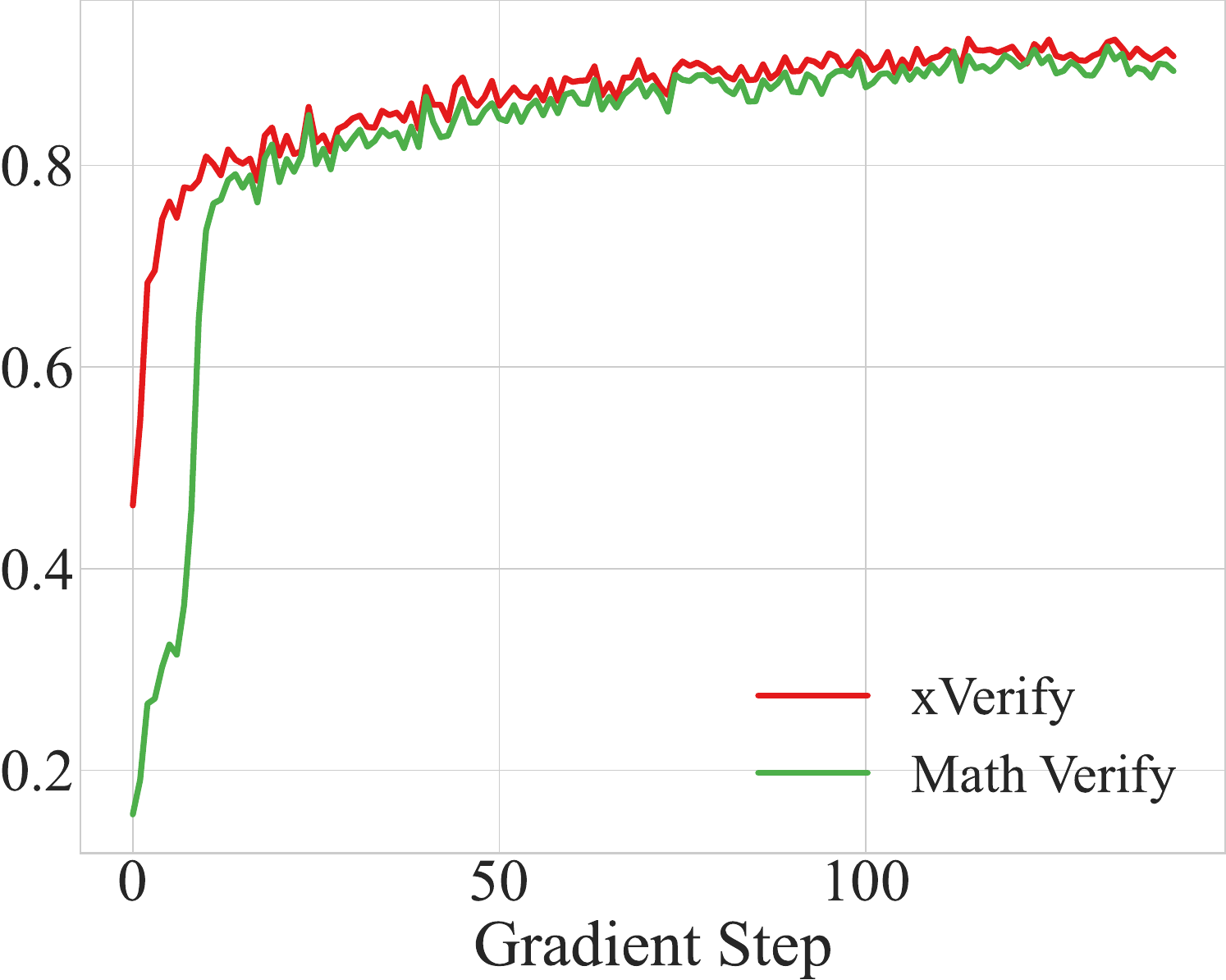}
    \caption{Learning Curves of Qwen2.5-7B in Reinforcement Learning}
    \label{fig:Learning Curves for Qwen2.5-7B}
\end{figure*}

\begin{figure*}[!ht]
    \centering
    \includegraphics[width=0.5\linewidth]{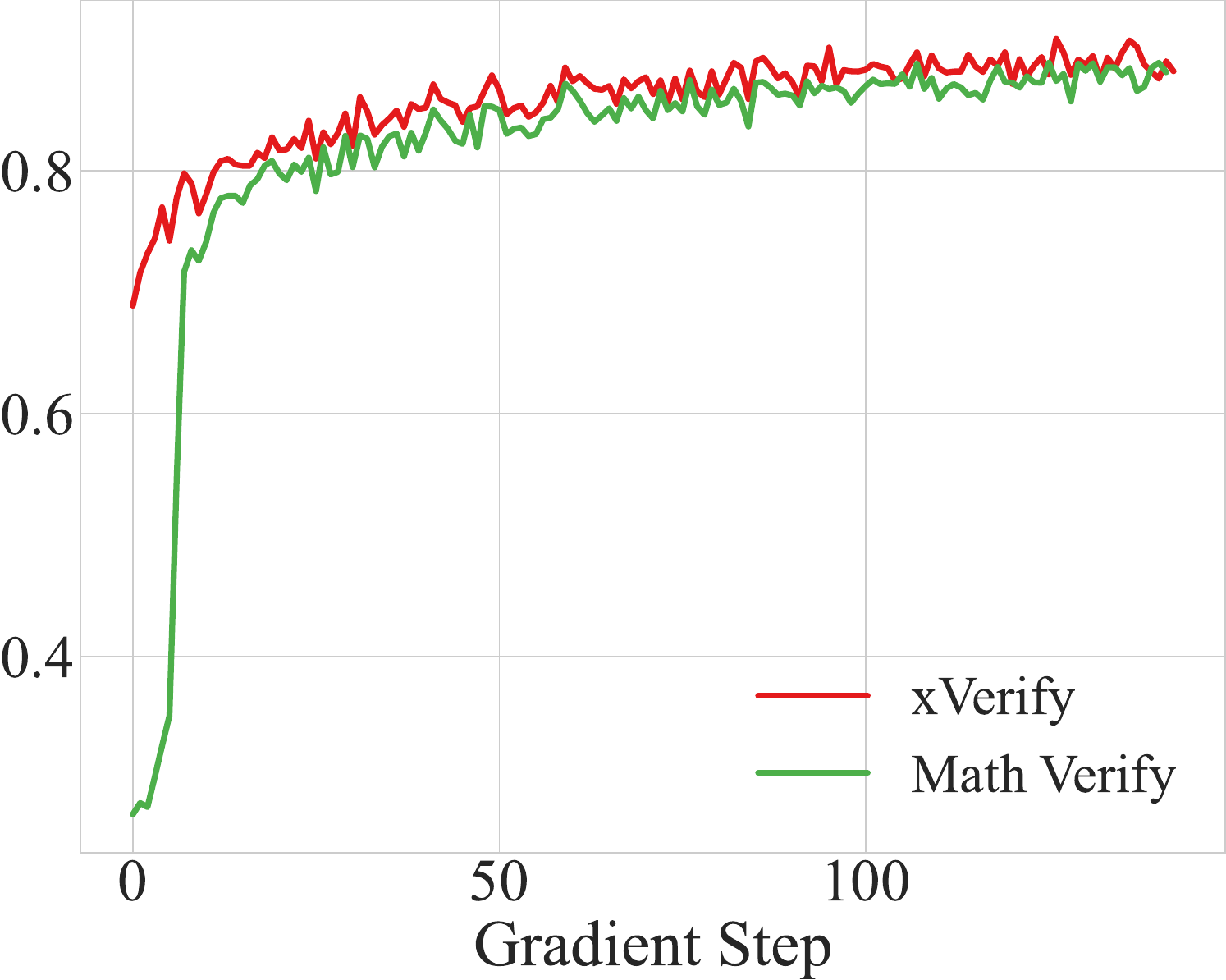}
    \caption{Learning Curves of Llama3.1-8B in Reinforcement Learning}
    \label{fig:Learning Curves for Llama3.1-8B}
\end{figure*}

Figures~\ref{fig:Learning Curves for Qwen2.5-7B} and~\ref{fig:Learning Curves for Llama3.1-8B} present the learning curves of Qwen2.5-7B and Llama3.1-8B during reinforcement learning training. It is evident from these curves that using xVerify as the reward model results in higher initial accuracy, indicating that xVerify provides more precise reward signals at the beginning of training. Moreover, the learning curves with xVerify converge faster, demonstrating that it delivers more accurate and reliable feedback, thereby significantly improving training efficiency. This is particularly important for larger-scale and broader reinforcement learning fine-tuning, where both training efficiency and reward signal quality are critical.

\subsection{Full Fine-Tuning vs. QLoRA for xVerify Models}
\label{appendix:full finetune vs. qlora}

\begin{table*}[!htbp]
    \centering
    \resizebox{.9\linewidth}{!}{
    \begin{threeparttable}
        \begin{tabular}{@{}c c c c c c@{}}
        \toprule
        \multirow{2}{*}{\textbf{Model}} & \multirow{2}{*}{\textbf{Fine-tuning Type}} & \multicolumn{2}{c}{\textbf{Test Set}} & \multicolumn{2}{c}{\textbf{Generalization Set}} \\ 
        \cmidrule(lr){3-4} \cmidrule(lr){5-6}
        & & \textbf{F1} & \textbf{Acc.} & \textbf{F1} & \textbf{Acc.} \\ 
        \midrule
        \multirow{2}{*}{Qwen2.5-0.5B-Instruct} & QLoRA & 96.69\% & 96.85\% & 95.29\% & 95.53\% \\
         & Full & 95.90\% & 96.10\% & 92.64\% & 93.03\% \\ 
        \midrule
        \multirow{2}{*}{LLaMA-3.2-3B-Instruct} & QLoRA & 97.02\% & 97.14\% & 96.11\% & 96.27\% \\
         & Full & 95.44\% & 95.67\% & 93.10\% & 93.41\% \\ 
        \bottomrule
        \end{tabular}
    \end{threeparttable}
    }
    \caption{Performance comparison of QLoRA and full fine-tuning on xVerify models.}
    \label{tab:model_perf}
\end{table*}

To quickly explore the impact of different fine-tuning strategies on xVerify models, we conducted a small-scale experiment comparing full fine-tuning with QLoRA on two representative models: Qwen2.5-0.5B-Instruct and LLaMA-3.2-3B-Instruct. Both models were fine-tuned on the VAR dataset, and their performance was evaluated on the test set as well as a held-out generalization set.

Table~\ref{tab:model_perf} summarizes the results. As shown, full fine-tuning achieves slightly lower overall performance compared to QLoRA. The decline is more pronounced on the generalization set, suggesting that full fine-tuning may be more prone to overfitting. In contrast, QLoRA maintains a better balance between test performance and generalization capability.

These findings indicate that, although full fine-tuning could in principle offer marginal gains under some conditions, in this small-scale exploration it does not outperform QLoRA. Additionally, QLoRA provides substantial computational savings, making it a more efficient choice for training instruction-tuned models. Overall, these results highlight that QLoRA is a practical and effective fine-tuning strategy for xVerify.

\subsection{Comparison of xVerify models with their base models}

\begin{table*}[!htbp]
    \centering
    \resizebox{.7\linewidth}{!}{
    \begin{threeparttable}
        \begin{tabular}{@{}c c c c c@{}}
        \toprule
        \multirow{2}{*}{\textbf{Model}} & \multicolumn{2}{c}{\textbf{Test Set}} & \multicolumn{2}{c}{\textbf{Generalization Set}} \\ 
        \cmidrule(lr){2-3} \cmidrule(lr){4-5}
         & \textbf{F1} & \textbf{Acc.} & \textbf{F1} & \textbf{Acc.} \\  
        \midrule
        Qwen2.5-0.5B-Instruct & 70.55\% & 66.71\% & 72.38\% & 68.78\% \\
        xVerify-0.5B-I        & \textbf{96.69\%} & \textbf{96.85\%} & \textbf{95.29\%} & \textbf{95.53\%} \\
        \midrule
        LLaMA-3.2-1B-Instruct & 37.15\% & 53.95\% & 41.87\% & 56.09\% \\
        xVerify-1B-I          & \textbf{96.77\%} & \textbf{96.91\%} & \textbf{95.43\%} & \textbf{95.62\%} \\
        \midrule
        LLaMA-3.2-3B-Instruct & 85.22\% & 84.99\% & 83.79\% & 83.32\% \\
        xVerify-3B-Ia         & \textbf{97.02\%} & \textbf{97.14\%} & \textbf{96.11\%} & \textbf{96.27\%} \\
        \midrule
        Gemma-2-9B-it         & 83.31\% & 81.89\% & 82.20\% & 80.61\% \\
        xVerify-9B-I          & \textbf{97.19\%} & \textbf{97.29\%} & \textbf{96.23\%} & \textbf{96.38\%} \\
        \bottomrule
        \end{tabular}
    \end{threeparttable}
    }
    \caption{Comparison of xVerify models with their base models on the test and generalization sets.}
    \label{tab:xverify_vs_base}
\end{table*}

To assess the effectiveness of VAR-based fine-tuning, we conducted a systematic evaluation of four xVerify variants and their corresponding base models on both the test and generalization sets, using identical evaluation settings. As shown in Table~\ref{tab:xverify_vs_base}, the fine-tuned xVerify models achieve over 10 percentage points of gain in both F1 and accuracy compared to their base models. Notably, xVerify-0.5B-I and xVerify-1B-I each deliver improvements of 20\%–50\% on these metrics. These results demonstrate that VAR-based fine-tuning substantially enhances the evaluation capabilities of the base models.

\subsection{Evaluation without the original question}
\label{appendix:eval without the original question}

\begin{table*}[!htbp]
    \centering
    \resizebox{.7\linewidth}{!}{
    \begin{threeparttable}
        \begin{tabular}{@{}l l l c c@{}}
        \toprule
        \textbf{Dataset} & \textbf{Model} & \textbf{Setting} & \textbf{F1} & \textbf{Acc.} \\ 
        \midrule
        \multirow{4}{*}{Test} 
            & \multirow{2}{*}{xVerify-0.5B-I} & with question & 96.69\% & 96.85\% \\
            &                                 & w/o question  & 96.53\% & 96.73\% \\ 
            \cmidrule(lr){2-5}
            & \multirow{2}{*}{xVerify-3B-Ia} & with question & 97.02\% & 97.14\% \\
            &                                & w/o question  & 96.09\% & 96.30\% \\
        \midrule
        \multirow{4}{*}{Generalization} 
            & \multirow{2}{*}{xVerify-0.5B-I} & with question & 95.29\% & 95.53\% \\
            &                                 & w/o question  & 95.05\% & 95.33\% \\ 
            \cmidrule(lr){2-5}
            & \multirow{2}{*}{xVerify-3B-Ia} & with question & 96.11\% & 96.27\% \\
            &                                & w/o question  & 95.55\% & 95.79\% \\
        \bottomrule
        \end{tabular}
    \end{threeparttable}
    }
    \caption{Performance of xVerify models with and without access to the original question.}
    \label{tab:xverify_question}
\end{table*}

In real-world evaluation scenarios, it is often the case that only the LLM’s response and the reference answer are available, without access to the original question. This setting arises, for example, when xVerify is used as a reward model in reinforcement learning.  

To assess whether xVerify can maintain strong accuracy and generalization under these conditions, we evaluated \textbf{xVerify-0.5B-I} and \textbf{xVerify-3B-Ia} on both the test set and the generalization set, with and without providing the original questions. In the modified setup, we removed only the parts of the prompt referencing the question, while keeping all other configurations identical.  

As shown in Table~\ref{tab:xverify_question}, both models exhibit virtually no performance drop across datasets, demonstrating that xVerify retains robust decision-making capability and generalization even in the absence of the original question. This suggests that xVerify’s core judgment mechanism relies primarily on the equivalence between the model’s output and the reference answer, while the original question serves only as auxiliary context rather than essential input.

\end{document}

%% file: tables/test_results.tex
\begin{table*}[!htbp]
    \vspace{-1.0em}
    \centering
    \resizebox{0.9\linewidth}{!}{
    \begin{threeparttable}
        \begin{tabular}{clcccccccccc}
        \toprule
        \multirow{2}{*}{\textbf{Method Type}} & \multicolumn{1}{c}{\multirow{2}{*}{\textbf{Method}}} & \multicolumn{2}{c}{\textbf{Multiple Choice}} & \multicolumn{2}{c}{\textbf{Math}} & \multicolumn{2}{c}{\textbf{Short Answer}} & \multicolumn{2}{c}{\textbf{Classification}} & \multicolumn{2}{c}{\textbf{Overall}} \\
        \cmidrule(lr){3-4} \cmidrule(lr){5-6} \cmidrule(lr){7-8} \cmidrule(lr){9-10} \cmidrule(lr){11-12}
         & \multicolumn{1}{c}{} & \textbf{F1} & \textbf{Acc.} & \textbf{F1} & \textbf{Acc.} & \textbf{F1} & \textbf{Acc.} & \textbf{F1} & \textbf{Acc.} & \textbf{F1} & \textbf{Acc.} \\
        \midrule 
        \multirow{6}{*}{\textbf{\begin{tabular}[c]{@{}c@{}}Evaluation \\ Framework\end{tabular}}} 
         & \textbf{DeepSeek Math Verify} & 70.77\% & 75.17\% & 78.34\% & 84.30\% & - & - & - & - & 74.90\% & 52.52\% \\
         & \textbf{LM Eval Harness} & 58.44\% & 68.19\% & 25.16\% & 28.27\% & 53.41\% & 44.51\% & 72.35\% & 66.94\% & 47.67\% & 48.32\% \\
         & \textbf{Math-Verify} & 5.88\% & 53.76\% & 82.55\% & 86.70\% & 42.27\% & 71.91\% & 0.00\% & 29.66\% & 45.64\% & 65.91\% \\
         & \textbf{OpenAI Simple Evals} & 23.61\% & 28.02\% & 66.79\% & 76.88\% & 42.23\% & 55.32\% & 73.29\% & 67.87\% & 51.17\% & 58.10\% \\
         & \textbf{OpenCompass} & 68.11\% & 72.52\% & 79.25\% & 84.73\% & - & - & - & - & 74.18\% & 79.64\% \\
         & \textbf{UltraEval} & 17.34\% & 18.04\% & 8.88\% & 56.89\% & - & - & - & - & 13.95\% & 40.71\% \\
        \midrule 
        \multirow{12}{*}{\textbf{\begin{tabular}[c]{@{}c@{}}Judge \\ Model\end{tabular}}} & \textbf{PandaLM-7B-v1} & 4.26\% & 8.12\% & 16.78\% & 14.46\% & 23.47\% & 17.72\% & 25.32\% & 16.79\% & 16.40\% & 13.72\% \\
         & \textbf{Auto-J-13B} & 40.00\% & 63.20\% & 26.32\% & 60.62\% & 64.41\% & 78.22\% & 86.04\% & 82.60\% & 53.38\% & 68.13\% \\
         & \textbf{Prometheus-8x7B-v2.0} & 71.26\% & 68.61\% & 71.99\% & 66.92\% & 76.24\% & 77.70\% & 83.27\% & 77.65\% & 74.57\% & 71.12\% \\
         & \textbf{JudgeLM-13B-v1.0} & 56.81\% & 48.89\% & 58.39\% & 59.46\% & 77.32\% & 79.52\% & 95.63\% & 93.82\% & 68.57\% & 65.83\% \\
         & \textbf{JudgeLM-33B-v1.0} & 42.86\% & 43.24\% & 44.82\% & 46.03\% & 57.86\% & 62.23\% & 73.42\% & 67.56\% & 52.00\% & 51.75\% \\
         & \textbf{CompassJudger-1-14B} & 58.94\% & 44.62\% & 55.09\% & 40.76\% & 59.66\% & 52.90\% & 90.87\% & 86.61\% & 63.22\% & 51.37\% \\
         & \textbf{CompassJudger-1-32B} & 95.09\% & 95.37\% & 84.11\% & 84.30\% & 94.95\% & 96.11\% & 98.45\% & 97.84\% & 91.67\% & 91.69\% \\
         & \textbf{GPT-4o as Judge} & 96.61\% & 96.75\% & 95.27\% & 95.80\% & 95.01\% & 96.20\% & 98.14\% & 97.43\% & 96.25\% & 96.39\% \\
         & \textbf{GPT-4o as Judge (CoT)} & 97.10\% & 97.23\% & 95.41\% & 95.88\% & 95.63\% & 96.63\% & 99.56\% & 99.38\% & 96.85\% & 96.95\% \\
        \midrule
        \rowcolor{gray!10}
         & \textbf{xVerify-0.5B-I} & \underline{97.78\%} & \textbf{97.90\%} & 93.74\% & 94.64\% & \textbf{96.72\%} & \textbf{97.49\%} & \underline{99.71\%} & \underline{99.59\%} & 96.69\% & 96.85\% \\
        \rowcolor{gray!10}
         & \textbf{xVerify-3B-Ib} & 97.31\% & 97.41\% & 95.65\% & 96.18\% & 96.38\% & 97.23\% & \textbf{99.78\%} & \textbf{99.69\%} & 97.17\% & 97.27\% \\
         \rowcolor{gray!10}
         & \textbf{xVerify-7B-I} & 97.75\% & \underline{97.84\%} & \textbf{95.94\%} & \textbf{96.44\%} & \underline{96.51\%} & \underline{97.32\%} & \textbf{99.78\%} & \textbf{99.69\%} & \textbf{97.41\%} & \textbf{97.50\%} \\
         \rowcolor{gray!10}
         \multirow{1}{*}[4pt]{\textbf{xVerify}} & \textbf{xVerify-9B-I} & 97.43\% & 97.53\% & 95.75\% & 96.27\% & 96.06\% & 96.97\% & \textbf{99.78\%} & \textbf{99.69\%} & 97.19\% & 97.29\% \\
         \rowcolor{gray!10}
         & \textbf{xVerify-14B-Ia} & 97.49\% & 97.59\% & 95.73\% & 96.22\% & 95.41\% & 96.46\% & 99.63\% & 99.49\% & 97.06\% & 97.16\% \\
         \rowcolor{gray!10}
         & \textbf{xVerify-32B-I} & \textbf{97.81\%} & \textbf{97.90\%} & \underline{95.88\%} & \underline{96.31\%} & 96.18\% & 97.06\% & \underline{99.71\%} & \underline{99.59\%} & \underline{97.32\%} & \underline{97.40\%} \\
         \bottomrule
        \end{tabular}
    \end{threeparttable}
    }
    \caption{Evaluation Results on the Test Set. "-" indicates that the evaluation method is not applicable to the problem type. Best and second-best results are \textbf{bold} and \underline{underlined}, respectively}
    \label{tab:test results}
    \vspace{-1.0em}
\end{table*}

%% file: tables/generalization_results.tex
\begin{table*}[!htbp]
    \vspace{-1.0em}
    \centering
    \resizebox{0.9\linewidth}{!}{
    \begin{threeparttable}
        \begin{tabular}{clcccccccccc}
        \toprule
        \multirow{2}{*}{\textbf{Method Type}} & \multicolumn{1}{c}{\multirow{2}{*}{\textbf{Method}}} & \multicolumn{2}{c}{\textbf{Multiple Choice}} & \multicolumn{2}{c}{\textbf{Math}} & \multicolumn{2}{c}{\textbf{Short Answer}} & \multicolumn{2}{c}{\textbf{Classification}} & \multicolumn{2}{c}{\textbf{Overall}} \\
        \cmidrule(lr){3-4} \cmidrule(lr){5-6} \cmidrule(lr){7-8} \cmidrule(lr){9-10} \cmidrule(lr){11-12}
         & \multicolumn{1}{c}{} & \textbf{F1} & \textbf{Acc.} & \textbf{F1} & \textbf{Acc.} & \textbf{F1} & \textbf{Acc.} & \textbf{F1} & \textbf{Acc.} & \textbf{F1} & \textbf{Acc.} \\
        \midrule
        \multirow{6}{*}{\textbf{\begin{tabular}[c]{@{}c@{}}Evaluation \\ Framework\end{tabular}}} 
         & \textbf{DeepSeek Math Verify} & 72.90\% & 73.39\% & 11.69\% & 79.83\% & - & - & - & - & 60.57\% & 44.42\% \\
         & \textbf{LM Eval Harness} & 61.60\% & 65.37\% & 7.03\% & 18.48\% & 58.22\% & 45.09\% & 92.06\% & 88.21\% & 55.81\% & 51.30\% \\
         & \textbf{Math-Verify} & 5.19\% & 45.10\% & 64.18\% & 87.68\% & 9.12\% & 52.75\% & 0.00\% & 24.59\% & 16.10\% & 55.53\% \\
         & \textbf{OpenAI Simple Evals} & 28.72\% & 29.23\% & 24.31\% & 78.90\% & 58.33\% & 59.58\% & 94.39\% & 91.62\% & 57.99\% & 63.36\% \\
         & \textbf{OpenCompass} & 71.64\% & 71.44\% & 47.22\% & 84.39\% & - & - & - & - & 65.74\% & 78.18\% \\
         & \textbf{UltraEval} & 16.29\% & 15.31\% & 13.55\% & 78.39\% & - & - & - & - & 15.71\% & 48.13\% \\
        \midrule
        \multirow{12}{*}{\textbf{\begin{tabular}[c]{@{}c@{}}Judge\\ Model\end{tabular}}} & \textbf{PandaLM-7B-v1} & 4.28\% & 7.85\% & 9.91\% & 15.97\% & 45.81\% & 31.43\% & 36.23\% & 25.99\% & 23.74\% & 19.14\% \\
         & \textbf{Auto-J-13B} & 34.87\% & 52.78\% & 9.86\% & 76.54\% & 85.12\% & 86.97\% & 77.67\% & 71.99\% & 60.43\% & 71.35\% \\
         & \textbf{Prometheus-8x7B-v2.0} & 74.13\% & 68.60\% & 49.48\% & 60.27\% & 87.15\% & 86.13\% & 84.70\% & 77.19\% & 74.51\% & 71.69\% \\
         & \textbf{JudgeLM-13B-v1.0} & 65.39\% & 57.80\% & 21.61\% & 44.87\% & 86.11\% & 84.53\% & 91.78\% & 86.89\% & 69.18\% & 65.63\% \\
         & \textbf{JudgeLM-33B-v1.0} & 46.99\% & 45.10\% & 20.31\% & 39.99\% & 71.34\% & 66.69\% & 41.92\% & 33.36\% & 46.06\% & 46.01\% \\
         & \textbf{CompassJudger-1-14B} & 63.65\% & 49.50\% & 27.63\% & 21.20\% & 73.61\% & 66.48\% & 88.97\% & 81.92\% & 63.10\% & 51.21\% \\
         & \textbf{CompassJudger-1-32B} & 92.93\% & 92.32\% & 72.05\% & 84.91\% & 96.81\% & 96.86\% & 98.05\% & 97.05\% & 91.90\% & 92.04\% \\
         & \textbf{GPT-4o as Judge} & 95.86\% & 95.38\% & 87.91\% & 94.76\% & 97.46\% & 97.49\% & 98.67\% & 97.98\% & 96.03\% & 96.18\% \\
         & \textbf{GPT-4o as Judge (CoT)} & 95.44\% & 94.88\% & 88.34\% & 94.71\% & 97.39\% & 97.42\% & 98.36\% & 97.52\% & 95.79\% & 95.92\% \\
        \midrule
        \rowcolor{gray!10}
        & \textbf{xVerify-0.5B-I} & \textbf{96.49\%} & \textbf{96.10\%} & 80.00\% & 91.94\% & 96.95\% & 97.00\% & \underline{99.03\%} & \underline{98.53\%} & 95.29\% & 95.53\% \\
        \rowcolor{gray!10}
         & \textbf{xVerify-3B-Ib} & 96.21\% & \underline{95.71\%} & 86.20\% & 94.15\% & \textbf{97.60\%} & \textbf{97.63\%} & \underline{99.03\%} & \underline{98.53\%} & 96.08\% & 96.23\% \\
         \rowcolor{gray!10}
         & \textbf{xVerify-7B-I} & 96.16\% & 95.66\% & 87.86\% & 94.87\% & 97.45\% & 97.49\% & 98.93\% & 98.37\% & 96.22\% & 96.37\% \\
         \rowcolor{gray!10}
         \multirow{1}{*}[4pt]{\textbf{xVerify}} & \textbf{xVerify-9B-I} & 96.06\% & 95.55\% & 87.47\% & 94.76\% & \underline{97.53\%} & \underline{97.56\%} & \textbf{99.13\%} & \textbf{98.68\%} & 96.23\% & 96.38\% \\
         \rowcolor{gray!10}
         & \textbf{xVerify-14B-Ia} & 96.11\% & 95.60\% & \textbf{90.20\%} & \textbf{95.74\%} & 97.32\% & 97.35\% & \textbf{99.13\%} & \textbf{98.68\%} & \textbf{96.53\%} & \textbf{96.65\%} \\
         \rowcolor{gray!10}
         & \textbf{xVerify-32B-I} & \underline{96.22\%} & \underline{95.71\%} & \underline{90.09\%} & \underline{95.59\%} & 97.32\% & 97.35\% & \underline{99.03\%} & \underline{98.53\%} & \underline{96.50\%} & \underline{96.60\%} \\
        \bottomrule
        \end{tabular}
    \end{threeparttable}
    }
    \caption{Evaluation Results on the Generalization Set. "-" indicates that the evaluation method is not applicable to the problem type. Best and second-best results are \textbf{bold} and \underline{underlined}, respectively}
    \label{tab:general results}
    \vspace{-0.5em}
\end{table*}

%% file: tables/rl_results.tex
\begin{table*}[htbp!]
  \centering
  \resizebox{0.8\linewidth}{!}{
  \begin{tabular}{lcccccccc}
    \toprule
    \textbf{Model} & MMLU-Pro  & GPQA  & MATH-500  & DROP &Amazon & CMMLU &CHID& Avg. \\
    \midrule
    \multicolumn{9}{c}{\textbf{Qwen2.5-7B}} \\
    Direct Generation & 40.4\% & 18.8\% & 39.6\% & 70.4\% &94.0\% &76.8\% &42.4\% &54.6\%\\
    \quad $\hookrightarrow$ RL with Math Verify & 52.0\% & \textbf{41.2\%} & 79.5\% & 84.8\% & 96.0\%&88.0\%& \textbf{64.0\%}&72.2\%\\
    \rowcolor{gray!10} \quad $\hookrightarrow$ RL with xVerify & \textbf{53.6\%} & 39.6\% & \textbf{81.5\%} & \textbf{86.0\%} & \textbf{98.4\%} & \textbf{88.4\%} &63.6\%&\textbf{73.0\%}\\
    \midrule
    \multicolumn{9}{c}{\textbf{Llama3.1-8B}}  \\
    Direct Generation & 42.8\% & 28.8\% & 52.6\% & 82.4\% & 96.0\% & 55.2\% &42.4\%&57.2\% \\
    \quad $\hookrightarrow$ RL with Math Verify & \textbf{46.4\%} & 32.4\% & 57.9\% & \textbf{88.8\%} & 97.2\% & 58.0\% & 42.4\%&60.4\%\\
    \rowcolor{gray!10} \quad $\hookrightarrow$ RL with xVerify & \textbf{46.4\%}& \textbf{34.4\%} & \textbf{58.0\%} & 87.6\% & \textbf{97.6\%} & \textbf{59.6\%} & \textbf{44.8\%}&\textbf{61.2\%}\\
    \bottomrule
  \end{tabular}
  }
  \caption{Evaluation Accuracy Results of RL with xVerify as Reward Model.}
  \label{tab:evaluation_results}
  \vspace{-1.em}
\end{table*}

%% file: tables/appendix_all_models_test_results.tex
\begin{table*}[!ht]
    \centering
    \resizebox{\linewidth}{!}{
    \begin{threeparttable}
        \begin{tabular}{clcccccccccc}
        \toprule
        \multirow{2}{*}{\textbf{Method Type}} & \multicolumn{1}{c}{\multirow{2}{*}{\textbf{Method}}} & \multicolumn{2}{c}{\textbf{Multiple Choice}} & \multicolumn{2}{c}{\textbf{Math}} & \multicolumn{2}{c}{\textbf{Short Answer}} & \multicolumn{2}{c}{\textbf{Classification}} & \multicolumn{2}{c}{\textbf{Overall}} \\
        \cmidrule(lr){3-4} \cmidrule(lr){5-6} \cmidrule(lr){7-8} \cmidrule(lr){9-10} \cmidrule(lr){11-12}
         & \multicolumn{1}{c}{} & \textbf{F1} & \textbf{Acc.} & \textbf{F1} & \textbf{Acc.} & \textbf{F1} & \textbf{Acc.} & \textbf{F1} & \textbf{Acc.} & \textbf{F1} & \textbf{Acc.} \\
        \midrule 
        \multirow{12}{*}{\textbf{\begin{tabular}[c]{@{}c@{}}Judge \\ Model\end{tabular}}} & \textbf{PandaLM-7B-v1} & 4.26\% & 8.12\% & 16.78\% & 14.46\% & 23.47\% & 17.72\% & 25.32\% & 16.79\% & 16.40\% & 13.72\% \\
         & \textbf{Auto-J-Bilingual-6B} & 52.85\% & 67.71\% & 40.76\% & 65.21\% & 67.22\% & 79.60\% & 74.86\% & 71.37\% & 57.04\% & 69.59\% \\
         & \textbf{Auto-J-13B} & 40.00\% & 63.20\% & 26.32\% & 60.62\% & 64.41\% & 78.22\% & 86.04\% & 82.60\% & 53.38\% & 68.13\% \\
         & \textbf{Prometheus-7B-v2.0} & 75.76\% & 75.41\% & 74.20\% & 74.35\% & 70.95\% & 74.59\% & 84.80\% & 77.03\% & 76.50\% & 75.11\% \\
         & \textbf{Prometheus-8x7B-v2.0} & 71.26\% & 68.61\% & 71.99\% & 66.92\% & 76.24\% & 77.70\% & 83.27\% & 77.65\% & 74.57\% & 71.12\% \\
         & \textbf{JudgeLM-7B-v1.0} & 56.53\% & 42.57\% & 46.09\% & 34.58\% & 60.33\% & 50.56\% & 83.89\% & 73.22\% & 59.02\% & 45.90\% \\
         & \textbf{JudgeLM-13B-v1.0} & 56.81\% & 48.89\% & 58.39\% & 59.46\% & 77.32\% & 79.52\% & 95.63\% & 93.82\% & 68.57\% & 65.83\% \\
         & \textbf{JudgeLM-33B-v1.0} & 42.86\% & 43.24\% & 44.82\% & 46.03\% & 57.86\% & 62.23\% & 73.42\% & 67.56\% & 52.00\% & 51.75\% \\
         & \textbf{CompassJudger-1-1.5B} & 49.95\% & 35.54\% & 61.66\% & 48.78\% & 57.36\% & 46.93\% & 82.51\% & 70.96\% & 61.94\% & 48.35\% \\
         & \textbf{CompassJudger-1-7B} & 70.05\% & 62.78\% & 66.62\% & 58.86\% & 67.47\% & 65.08\% & 92.99\% & 89.50\% & 72.72\% & 65.96\% \\
         & \textbf{CompassJudger-1-14B} & 58.94\% & 44.62\% & 55.09\% & 40.76\% & 59.66\% & 52.90\% & 90.87\% & 86.61\% & 63.22\% & 51.37\% \\
         & \textbf{CompassJudger-1-32B} & 95.09\% & 95.37\% & 84.11\% & 84.30\% & 94.95\% & 96.11\% & 98.45\% & 97.84\% & 91.67\% & 91.69\% \\
         & \textbf{GPT-4o as Judge} & 96.61\% & 96.75\% & 95.27\% & 95.80\% & 95.01\% & 96.20\% & 98.14\% & 97.43\% & 96.25\% & 96.39\% \\
         & \textbf{GPT-4o as Judge (CoT)} & 97.10\% & 97.23\% & 95.41\% & 95.88\% & 95.63\% & 96.63\% & 99.56\% & 99.38\% & 96.85\% & 96.95\% \\
        \midrule
        \rowcolor{gray!10}
         & \textbf{xVerify-0.5B-I} & 97.78\% & 97.90\% & 93.74\% & 94.64\% & \textbf{96.72\%} & \textbf{97.49\%} & \underline{99.71\%} & \underline{99.59\%} & 96.69\% & 96.85\% \\
        \rowcolor{gray!10}
         & \textbf{xVerify-1B-I} & 97.22\% & 97.35\% & 94.76\% & 95.45\% & 96.06\% & 96.97\% & \underline{99.71\%} & \underline{99.59\%} & 96.77\% & 96.91\% \\
        \rowcolor{gray!10}
         & \textbf{xVerify-1.5B-I} & 97.85\% & 97.96\% & 95.10\% & 95.75\% & 96.05\% & 96.97\% & 99.63\% & 99.49\% & 97.05\% & 97.17\% \\
        \rowcolor{gray!10}
         & \textbf{xVerify-2B-I} & \underline{97.93\%} & \underline{98.02\%} & 95.06\% & 95.71\% & 96.06\% & 96.97\% & \textbf{99.78\%} & \textbf{99.69\%} & 97.09\% & 97.21\% \\
        \rowcolor{gray!10}
         & \textbf{xVerify-3B-Ia} & 97.73\% & 97.84\% & 95.00\% & 95.67\% & 96.17\% & 97.06\% & \underline{99.71\%} & \underline{99.59\%} & 97.02\% & 97.14\% \\
        \rowcolor{gray!10}
         & \textbf{xVerify-3B-Ib} & 97.31\% & 97.41\% & 95.65\% & 96.18\% & 96.38\% & 97.23\% & \textbf{99.78\%} & \textbf{99.69\%} & 97.17\% & 97.27\% \\
        \rowcolor{gray!10}
         & \textbf{xVerify-7B-I} & 97.75\% & 97.84\% & \underline{95.94\%} & \underline{96.44\%} & \underline{96.51\%} & \underline{97.32\%} & \textbf{99.78\%} & \textbf{99.69\%} & \textbf{97.41\%} & \textbf{97.50\%} \\
        \rowcolor{gray!10}
         \multirow{1}{*}[4pt]{\textbf{xVerify}} & \textbf{xVerify-8B-I} & 97.92\% & \underline{98.02\%} & 95.34\% & 95.97\% & 96.05\% & 96.97\% & \underline{99.71\%} & \underline{99.59\%} & 97.17\% & 97.29\% \\
        \rowcolor{gray!10}
         & \textbf{xVerify-9B-C} & \textbf{98.29\%} & \textbf{98.38\%} & 95.26\% & 95.88\% & 96.06\% & 96.97\% & \textbf{99.78\%} & \textbf{99.69\%} & 97.25\% & 97.37\% \\
        \rowcolor{gray!10}
         & \textbf{xVerify-9B-I} & 97.43\% & 97.53\% & 95.75\% & 96.27\% & 96.06\% & 96.97\% & \textbf{99.78\%} & \textbf{99.69\%} & 97.19\% & 97.29\% \\
        \rowcolor{gray!10}
         & \textbf{xVerify-14B-Ia} & 97.49\% & 97.59\% & 95.73\% & 96.22\% & 95.41\% & 96.46\% & 99.63\% & 99.49\% & 97.06\% & 97.16\% \\
        \rowcolor{gray!10}
         & \textbf{xVerify-14B-Ib} & 97.67\% & 97.78\% & \textbf{96.10\%} & \textbf{96.57\%} & 95.74\% & 96.72\% & \underline{99.71\%} & \underline{99.59\%} & 97.31\% & \underline{97.40\%} \\
        \rowcolor{gray!10}
         & \textbf{xVerify-27B-I} & 97.81\% & 97.90\% & 95.46\% & 96.01\% & 96.19\% & 97.06\% & 99.56\% & 99.38\% & 97.15\% & 97.26\% \\
        \rowcolor{gray!10}
         & \textbf{xVerify-32B-I} & 97.81\% & 97.90\% & 95.88\% & 96.31\% & 96.18\% & 97.06\% & \underline{99.71\%} & \underline{99.59\%} & \underline{97.32\%} & \underline{97.40\%} \\
         \bottomrule
        \end{tabular}
    \end{threeparttable}
    }
    \caption{Evaluation Accuracy Results on the Test Set: All xVerify Models and Judge Models. The best performance in each column is shown in \textbf{bold}, and the second-best performance is \underline{underlined}.}
    \label{tab:appendix test results}
\end{table*}

%% file: tables/appendix_all_models_generalization_results.tex
\begin{table*}[!htbp]
    \centering
    \resizebox{\linewidth}{!}{
    \begin{threeparttable}
        \begin{tabular}{clcccccccccc}
        \toprule
        \multirow{2}{*}{\textbf{Method Type}} & \multicolumn{1}{c}{\multirow{2}{*}{\textbf{Method}}} & \multicolumn{2}{c}{\textbf{Multiple Choice}} & \multicolumn{2}{c}{\textbf{Math}} & \multicolumn{2}{c}{\textbf{Short Answer}} & \multicolumn{2}{c}{\textbf{Classification}} & \multicolumn{2}{c}{\textbf{Overall}} \\
        \cmidrule(lr){3-4} \cmidrule(lr){5-6} \cmidrule(lr){7-8} \cmidrule(lr){9-10} \cmidrule(lr){11-12}
         & \multicolumn{1}{c}{} & \textbf{F1} & \textbf{Acc.} & \textbf{F1} & \textbf{Acc.} & \textbf{F1} & \textbf{Acc.} & \textbf{F1} & \textbf{Acc.} & \textbf{F1} & \textbf{Acc.} \\
        \midrule
        \multirow{12}{*}{\textbf{\begin{tabular}[c]{@{}c@{}}Judge\\ Model\end{tabular}}} & \textbf{PandaLM-7B-v1} & 4.28\% & 7.85\% & 9.91\% & 15.97\% & 45.81\% & 31.43\% & 36.23\% & 25.99\% & 23.74\% & 19.14\% \\
         & \textbf{Auto-J-Bilingual-6B} & 52.07\% & 60.75\% & 10.56\% & 74.79\% & 85.16\% & 86.76\% & 84.90\% & 79.91\% & 67.20\% & 74.57\% \\
         & \textbf{Auto-J-13B} & 34.87\% & 52.78\% & 9.86\% & 76.54\% & 85.12\% & 86.97\% & 77.67\% & 71.99\% & 60.43\% & 71.35\% \\
         & \textbf{Prometheus-7B-v2.0} & 76.67\% & 73.66\% & 49.08\% & 71.46\% & 81.52\% & 81.32\% & 79.59\% & 71.92\% & 73.85\% & 74.35\% \\
         & \textbf{Prometheus-8x7B-v2.0} & 74.13\% & 68.60\% & 49.48\% & 60.27\% & 87.15\% & 86.13\% & 84.70\% & 77.19\% & 74.51\% & 71.69\% \\
         & \textbf{JudgeLM-7B-v1.0} & 60.22\% & 45.71\% & 12.71\% & 15.40\% & 72.15\% & 62.51\% & 86.11\% & 76.18\% & 59.11\% & 46.38\% \\
         & \textbf{JudgeLM-13B-v1.0} & 65.39\% & 57.80\% & 21.61\% & 44.87\% & 86.11\% & 84.53\% & 91.78\% & 86.89\% & 69.18\% & 65.63\% \\
         & \textbf{JudgeLM-33B-v1.0} & 46.99\% & 45.10\% & 20.31\% & 39.99\% & 71.34\% & 66.69\% & 41.92\% & 33.36\% & 46.06\% & 46.01\% \\
         & \textbf{CompassJudger-1-1.5B} & 55.75\% & 40.87\% & 34.53\% & 33.62\% & 63.93\% & 51.57\% & 84.49\% & 73.93\% & 60.01\% & 47.65\% \\
         & \textbf{CompassJudger-1-7B} & 74.31\% & 65.20\% & 38.27\% & 39.89\% & 88.99\% & 88.15\% & 93.29\% & 89.29\% & 73.47\% & 67.47\% \\
         & \textbf{CompassJudger-1-14B} & 63.65\% & 49.50\% & 27.63\% & 21.20\% & 73.61\% & 66.48\% & 88.97\% & 81.92\% & 63.10\% & 51.21\% \\
         & \textbf{CompassJudger-1-32B} & 92.93\% & 92.32\% & 72.05\% & 84.91\% & 96.81\% & 96.86\% & 98.05\% & 97.05\% & 91.90\% & 92.04\% \\
         & \textbf{GPT-4o as Judge} & 95.86\% & 95.38\% & 87.91\% & 94.76\% & 97.46\% & 97.49\% & 98.67\% & 97.98\% & 96.03\% & 96.18\% \\
         & \textbf{GPT-4o as Judge (CoT)} & 95.44\% & 94.88\% & 88.34\% & 94.71\% & 97.39\% & 97.42\% & 98.36\% & 97.52\% & 95.79\% & 95.92\% \\
        \midrule
        \rowcolor{gray!10}
         & \textbf{xVerify-0.5B-I} & 96.49\% & 96.10\% & 80.00\% & 91.94\% & 96.95\% & 97.00\% & \underline{99.03\%} & \underline{98.53\%} & 95.29\% & 95.53\% \\
        \rowcolor{gray!10}
         & \textbf{xVerify-1B-I} & 96.10\% & 95.66\% & 82.45\% & 92.51\% & 97.32\% & 97.35\% & 98.92\% & 98.37\% & 95.43\% & 95.62\% \\
        \rowcolor{gray!10}
         & \textbf{xVerify-1.5B-I} & \underline{96.76\%} & \underline{96.38\%} & 83.58\% & 93.12\% & 97.46\% & 97.49\% & 98.88\% & 98.29\% & 95.85\% & 96.03\% \\
        \rowcolor{gray!10}
         & \textbf{xVerify-2B-I} & 96.27\% & 95.82\% & 82.11\% & 92.51\% & \textbf{97.60\%} & \textbf{97.63\%} & 98.98\% & 98.45\% & 95.57\% & 95.75\% \\
        \rowcolor{gray!10}
         & \textbf{xVerify-3B-Ia} & 96.44\% & 95.99\% & 86.10\% & 94.25\% & 97.31\% & 97.35\% & \underline{99.03\%} & \underline{98.53\%} & 96.11\% & 96.27\% \\
        \rowcolor{gray!10}
         & \textbf{xVerify-3B-Ib} & 96.21\% & 95.71\% & 86.20\% & 94.15\% & \textbf{97.60\%} & \textbf{97.63\%} & \underline{99.03\%} & \underline{98.53\%} & 96.08\% & 96.23\% \\
        \rowcolor{gray!10}
         & \textbf{xVerify-7B-I} & 96.16\% & 95.66\% & 87.86\% & 94.87\% & 97.45\% & 97.49\% & 98.93\% & 98.37\% & 96.22\% & 96.37\% \\
        \rowcolor{gray!10}
        \multirow{1}{*}[4pt]{\textbf{xVerify}} & \textbf{xVerify-8B-I} & 96.67\% & 96.27\% & 86.76\% & 94.61\% & 97.45\% & 97.49\% & \underline{99.03\%} & \underline{98.53\%} & 96.33\% & 96.49\% \\
        \rowcolor{gray!10}
         & \textbf{xVerify-9B-C} & \textbf{97.00\%} & \textbf{96.66\%} & 87.08\% & 94.71\% & 97.45\% & 97.49\% & 98.98\% & 98.45\% & 96.45\% & \underline{96.61\%} \\
        \rowcolor{gray!10}
         & \textbf{xVerify-9B-I} & 96.06\% & 95.55\% & 87.47\% & 94.76\% & \underline{97.53\%} & \underline{97.56\%} & \textbf{99.13\%} & \textbf{98.68\%} & 96.23\% & 96.38\% \\
        \rowcolor{gray!10}
         & \textbf{xVerify-14B-Ia} & 96.11\% & 95.60\% & \textbf{90.20\%} & \textbf{95.74\%} & 97.32\% & 97.35\% & \textbf{99.13\%} & \textbf{98.68\%} & \textbf{96.53\%} & \textbf{96.65\%} \\
        \rowcolor{gray!10}
         & \textbf{xVerify-14B-Ib} & 96.35\% & 95.88\% & 87.88\% & 94.92\% & 97.45\% & 97.49\% & 98.93\% & 98.37\% & 96.30\% & 96.44\% \\
        \rowcolor{gray!10}
         & \textbf{xVerify-27B-I} & 96.01\% & 95.49\% & 85.64\% & 93.99\% & 97.32\% & 97.35\% & \textbf{99.13\%} & \textbf{98.68\%} & 95.93\% & 96.09\% \\
        \rowcolor{gray!10}
         & \textbf{xVerify-32B-I} & 96.22\% & 95.71\% & \underline{90.09\%} & \underline{95.59\%} & 97.32\% & 97.35\% & \underline{99.03\%} & \underline{98.53\%} & \underline{96.50\%} & 96.60\% \\
        \bottomrule
        \end{tabular}
    \end{threeparttable}
    }
    \caption{Evaluation Accuracy Results on the Generalization Set: All xVerify Models and Judge Models. The best performance in each column is shown in \textbf{bold}, and the second-best performance is \underline{underlined}.}
    \label{tab:appendix general results}
\end{table*}

%% file: tables/appendix_time_costs.tex
\begin{table*}[!ht]
    \centering
    \resizebox{\linewidth}{!}{
    \begin{threeparttable}
        \begin{tabular}{clccccc}
        \toprule
        \textbf{Method Type} & \multicolumn{1}{c}{\textbf{Method}} & \textbf{Multiple Choice (s)} & \textbf{Math (s)} & \textbf{Short Answer (s)} & \textbf{Classification (s)} & \textbf{Avg (s)} \\
        \midrule
        \multirow{12}{*}{\textbf{\begin{tabular}[c]{@{}c@{}}Judge\\ Model\end{tabular}}} & \textbf{PandaLM-7B-v1} & 304.50 & 76.24 & 76.97 & 65.79 & 130.88 \\
         & \textbf{Auto-J-Bilingual-6B} & 1,570.44 & 1,802.71 & 1,194.08 & 1,148.32 & 1,428.89 \\
         & \textbf{Auto-J-13B} & 3,055.00 & 3,622.70 & 2,807.23 & 1,903.00 & 2,846.98 \\
         & \textbf{Prometheus-7B-v2.0} & 1,173.80 & 947.71 & 706.74 & 696.34 & 881.15 \\
         & \textbf{Prometheus-8x7B-v2.0} & 1,557.10 & 1,128.08 & 1,132.84 & 750.51 & 1,142.13 \\
         & \textbf{JudgeLM-7B-v1.0} & 551.88 & 469.10 & 394.57 & 348.05 & 440.90 \\
         & \textbf{JudgeLM-13B-v1.0} & 777.73 & 598.19 & 564.25 & 529.60 & 617.44 \\
         & \textbf{JudgeLM-33B-v1.0} & 1,041.83 & 1,018.37 & 789.80 & 762.99 & 903.25 \\
         & \textbf{CompassJudger-1-1.5B} & 189.45 & 244.08 & 139.50 & 110.95 & 171.00 \\
         & \textbf{CompassJudger-1-7B} & 163.96 & 568.72 & 450.20 & 80.58 & 315.87 \\
         & \textbf{CompassJudger-1-14B} & 346.80 & 571.66 & 217.86 & 196.18 & 333.13 \\
         & \textbf{CompassJudger-1-32B} & 147.53 & 258.10 & 133.59 & 152.11 & 172.83 \\
         \midrule
        \rowcolor{gray!10}
         & \textbf{xVerify-0.5B-I} & 38.97 & \underline{41.25} & 39.12 & 38.87 & \underline{39.55} \\
        \rowcolor{gray!10}
         & \textbf{xVerify-1B-I} & \textbf{33.91} & \textbf{36.63} & \textbf{33.44} & \textbf{33.47} & \textbf{34.36} \\
        \rowcolor{gray!10}
         & \textbf{xVerify-1.5B-I} & 43.05 & 46.87 & 42.17 & 42.08 & 43.54 \\
        \rowcolor{gray!10}
         & \textbf{xVerify-2B-I} & \underline{38.44} & 73.16 & 39.29 & \underline{37.38} & 47.07 \\
        \rowcolor{gray!10}
         & \textbf{xVerify-3B-Ia} & 38.54 & 44.54 & \underline{37.11} & 43.02 & 40.80 \\
        \rowcolor{gray!10}
         & \textbf{xVerify-3B-Ib} & 46.93 & 53.58 & 106.06 & 47.84 & 63.60 \\
        \rowcolor{gray!10}
         & \textbf{xVerify-7B-I} & 68.24 & 95.50 & 50.66 & 51.67 & 66.52 \\
        \rowcolor{gray!10}
        \multirow{1}{*}[4pt]{\textbf{xVerify}} & \textbf{xVerify-8B-I} & 78.06 & 61.57 & 45.34 & 46.82 & 57.95 \\
        \rowcolor{gray!10}
         & \textbf{xVerify-9B-C} & 131.07 & 70.16 & 51.66 & 52.57 & 76.37 \\
        \rowcolor{gray!10}
         & \textbf{xVerify-9B-I} & 54.20 & 69.91 & 49.41 & 51.06 & 56.15 \\
        \rowcolor{gray!10}
         & \textbf{xVerify-14B-Ia} & 59.18 & 114.91 & 55.50 & 54.80 & 71.10 \\
        \rowcolor{gray!10}
         & \textbf{xVerify-14B-Ib} & 61.17 & 145.19 & 116.43 & 57.55 & 95.09 \\
        \rowcolor{gray!10}
         & \textbf{xVerify-27B-I} & 85.28 & 89.41 & 58.99 & 61.00 & 73.67 \\
        \rowcolor{gray!10}
         & \textbf{xVerify-32B-I} & 131.05 & 98.99 & 64.74 & 67.45 & 90.56 \\
         \bottomrule
        \end{tabular}
    \end{threeparttable}
    }
    \caption{Running Time Comparison of xVerify Models and Other Judge Models (200 Samples per Question Type). The best performance in each column is shown in \textbf{bold}, and the second-best performance is \underline{underlined}.}
    \label{tab:appendix time cost results}
\end{table*}